\documentclass[10pt,twocolumn,letterpaper]{article}

\usepackage[pagenumbers]{cvpr} 

\usepackage{graphicx}
\usepackage{amsmath}
\usepackage{amssymb}
\usepackage{booktabs}

\usepackage{stackengine}
\usepackage{multirow}
\usepackage{tabularx}
\usepackage{wrapfig}
\usepackage{threeparttable}

\usepackage[pagebackref,breaklinks,colorlinks]{hyperref}

\usepackage[capitalize]{cleveref}
\crefname{section}{Sec.}{Secs.}
\Crefname{section}{Section}{Sections}
\Crefname{table}{Table}{Tables}
\crefname{table}{Tab.}{Tabs.}

\def\cvprPaperID{2995} 
\def\confName{CVPR}
\def\confYear{2022}

\begin{document}

\title{Local Texture Estimator for Implicit Representation Function}

\author{
Jaewon Lee\qquad Kyong Hwan Jin\thanks{Corresponding author.}\\
Daegu Gyeongbuk Institute of Science and Technology (DGIST), Korea\\
{\tt\small \{ljw3136, kyong.jin\}@dgist.ac.kr}\\
\url{https://github.com/jaewon-lee-b/lte}
}
\maketitle

\begin{abstract}
Recent works with an implicit neural function shed light on representing images in arbitrary resolution. However, a standalone multi-layer perceptron shows limited performance in learning high-frequency components. In this paper, we propose a Local Texture Estimator (LTE), a dominant-frequency estimator for natural images, enabling an implicit function to capture fine details while reconstructing images in a continuous manner. When jointly trained with a deep super-resolution (SR) architecture, LTE is capable of characterizing image textures in 2D Fourier space.
We show that an LTE-based neural function achieves favorable performance against existing deep SR methods within an arbitrary-scale factor.
Furthermore, we demonstrate that our implementation takes the shortest running time compared to previous works.
\end{abstract}
\vspace{-10pt}

\section{Introduction}
Single image super-resolution (SISR) is one of the most fundamental problems in computer vision and graphics. SISR aims to reconstruct high-resolution (HR) images from its degraded low-resolution (LR) counterpart. Dominant approaches \cite{DBLP:journals/corr/ShiCHTABRW16,Lim_2017_CVPR_Workshops, zhang2018rcan, zhang2018residual, dai2019second, Mei_2021_CVPR, DBLP:conf/cvpr/Chen000DLMX0021, liang2021swinir} are to extract feature maps using a deep vision architecture and then upsample to HR images at the end of a network. However, we need to train and store several models for each scale factor when an upsampler is implemented by sub-pixel convolution \cite{DBLP:journals/corr/ShiCHTABRW16}. In contrast, arbitrary-scale SR methods \cite{hu2019meta, chen2021learning} are promising since such ideas pave the way to restore images in a continuous manner with only a single network. 

Recently, implicit neural functions parameterized by a multi-layer perceptron (MLP) achieved remarkable performance in representing continuous-domain signals, such as images \cite{chen2021learning}, occupancy \cite{Occupancy_Networks}, signed distance \cite{Park_2019_CVPR}, shape representation \cite{Local_Implicit_Grid_CVPR20}, and view synthesis \cite{sitzmann2019srns,mildenhall2020nerf}. Such MLPs take coordinates as inputs and are trained in a framework of gradient descent optimization and machine learning. Inspired by recent progress in implicit representation, LIIF \cite{chen2021learning} replaced sub-pixel convolution with MLPs to accomplish arbitrary-scale SR, even at substantial scale factors.

One limitation of implicit neural representations is that a standalone MLP is biased towards learning low-frequency components \cite{DBLP:conf/icml/RahamanBADLHBC19} and fails to capture fine details \cite{tancik2020fourfeat}. Such phenomenon is referred to as spectral bias, and recent lines of research in resolving this problem are projecting input coordinates into a high-dimensional Fourier feature space \cite{mildenhall2020nerf, tancik2020fourfeat} or substituting a ReLU with a sinusoidal activation \cite{sitzmann2019siren}. Motivated from previous works, we study arbitrary-scale SISR problems through the lens of Fourier analysis.

\begin{figure}[t]
\centering
\includegraphics[scale = 0.4]{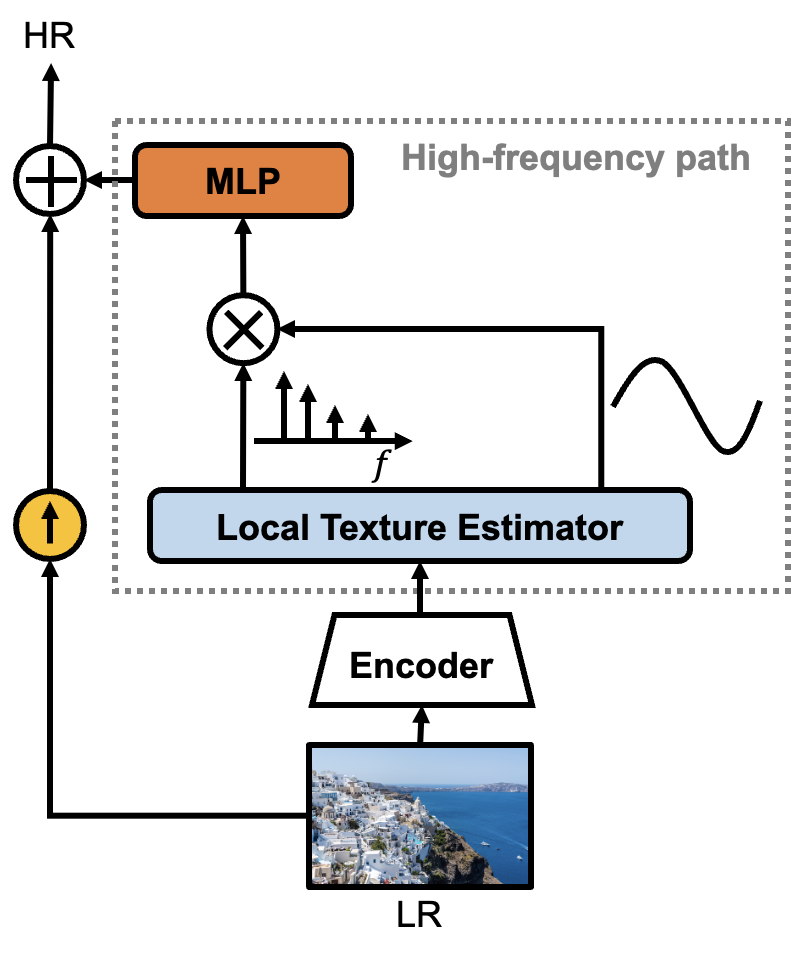}
\vspace{-10pt}
\caption{\textbf{Overview of Local Texture Estimator (LTE).} Our LTE estimates dominant frequencies and corresponding Fourier coefficients for natural images. Then, an MLP makes use of the estimated essential Fourier information to reconstruct an HR image in arbitrary resolution. We provide low-frequency information by adding an upscaled LR image to the output of MLP.}
\label{fig:concept}
\vspace{-10pt}
\end{figure}

In this paper, we propose a Local Texture Estimator (LTE), a dominant-frequency estimator for natural images, to allow an implicit function to learn fine details while restoring images in arbitrary resolution. We assume that an implicit function prioritizes learning image textures when utilizing dominant frequencies of images, as described in \cref{fig:concept}. Let us take an example from an image with vertical textures. Intuitively, the dominant frequencies of such an image are located on an $\mathbf x$-axis in 2D Fourier space. We observe that LTE is capable of extracting such dominant frequencies, characterizing image textures in 2D Fourier space, when jointly trained with a deep SR architecture, such as EDSR \cite{Lim_2017_CVPR_Workshops}, RDN \cite{zhang2018residual}, and SwinIR \cite{liang2021swinir}. In addition to extracting dominant frequencies, we show that estimating Fourier coefficients is also essential in improving a representational power of an implicit function in \cref{sec:abl}.

Short computation time is essential in SR applications. In addition, restoring larger than 2K-sized images is a memory-consuming task. Hence, we study the computation time of our LTE at various memory conditions and demonstrate that our approach is faster compared to previous works regardless of memory setting in \cref{sec:dis}.



In summary, our main contributions are as follows:
\begin{itemize}
    \item We point out that a deep SR network followed by LTE is capable of estimating dominant frequencies and corresponding Fourier coefficients for natural images.
    \item We show that an implicit representation function for arbitrary resolution prioritizes learning high-frequency details when essential Fourier information is estimated by LTE.
    \item We investigate the computation time of our LTE at several memory settings and demonstrate that our approach is faster compared to previous works.
\end{itemize}

\section{Related Work}
{\bf Implicit neural representation} Based on the fact that neural networks are universal function approximators \cite{10.5555/70405.70408}, coordinate-based MLP is widely applied to represent continuous-domain signals in computer vision and graphics. Such MLPs are a memory-efficient framework for HR data since the amount of memory to store MLPs is independent of data resolution. However, pre-trained MLPs show limited performance in representing unseen data, requiring training per each signal. To overcome this generalization issue, \cite{sitzmann2019srns, sitzmann2019siren} trained representation with a hypernetwork, which maps latent variables to weights of an MLP as in \cite{DBLP:conf/iclr/HaDL17}. In \cite{Park_2019_CVPR, Occupancy_Networks}, an MLP takes not only coordinates but also latent variables as inputs to enable representation to be a function of data. Recently, meta-learned initialization \cite{tancik2020meta} has been proposed, which provides a strong prior for representing signals, leading to fast convergence. Our work is mainly related to local implicit neural representation \cite{Local_Implicit_Grid_CVPR20, chen2021learning}. Such approaches assume that latent variables are evenly distributed over space, allowing an implicit function to focus on learning local features. Inspired by this, we hypothesize that the local region in HR representation shares the same image textures.

\textbf{Spectral bias} Recent works \cite{DBLP:conf/icml/RahamanBADLHBC19, sitzmann2019siren, mildenhall2020nerf, tancik2020fourfeat} have shown that a standard MLP with ReLUs shows limited performance in representing high-frequency textures. Such a phenomenon is referred to as spectral bias, and various methods have been proposed to alleviate this problem. Recently, SIREN \cite{sitzmann2019siren} used the sine layer as a non-linear activation instead of a ReLU, resulting in fast convergence and high data fidelity. Other approaches \cite{mildenhall2020nerf, tancik2020fourfeat} are to map input coordinates into high dimensional Fourier space by using position encoding or Fourier feature mapping before passing an MLP. Frequencies are fixed to the power of two \cite{mildenhall2020nerf} or randomly sampled from Gaussian distribution \cite{tancik2020fourfeat}. Unlike previous works, dominant frequencies from our LTE are data-driven and characterize high-frequency textures in 2D Fourier space.

{\bf Deep SR architecture} After ESPCN \cite{DBLP:journals/corr/ShiCHTABRW16} has proposed an efficient learnable upsampling module using pixel-shuffling, numerous CNN-based approaches \cite{Lim_2017_CVPR_Workshops, zhang2018rcan, zhang2018residual, dai2019second, Mei_2021_CVPR} have been studied to improve the representational power of models. Such approaches have exploited more complicated neural network architecture designs, such as residual block \cite{Lim_2017_CVPR_Workshops}, densely connected residual block \cite{zhang2018residual}, channel attention \cite{zhang2018rcan, dai2019second}, or non-local neural networks \cite{Mei_2021_CVPR}. Inspired by the success of the self-attention mechanism in high-level vision tasks \cite{DBLP:conf/iclr/DosovitskiyB0WZ21, liu2021swin}, general-purpose image processing transformers, such as IPT \cite{DBLP:conf/cvpr/Chen000DLMX0021} or SwinIR \cite{liang2021swinir}, have been proposed. Even though transformer-based SR architectures using a large dataset surpass CNN-based architectures in performance, these approaches nevertheless need to utilize a specific upscale module for each upsampling rate.

\begin{figure*}[t]
\centering
\includegraphics[scale = 0.4]{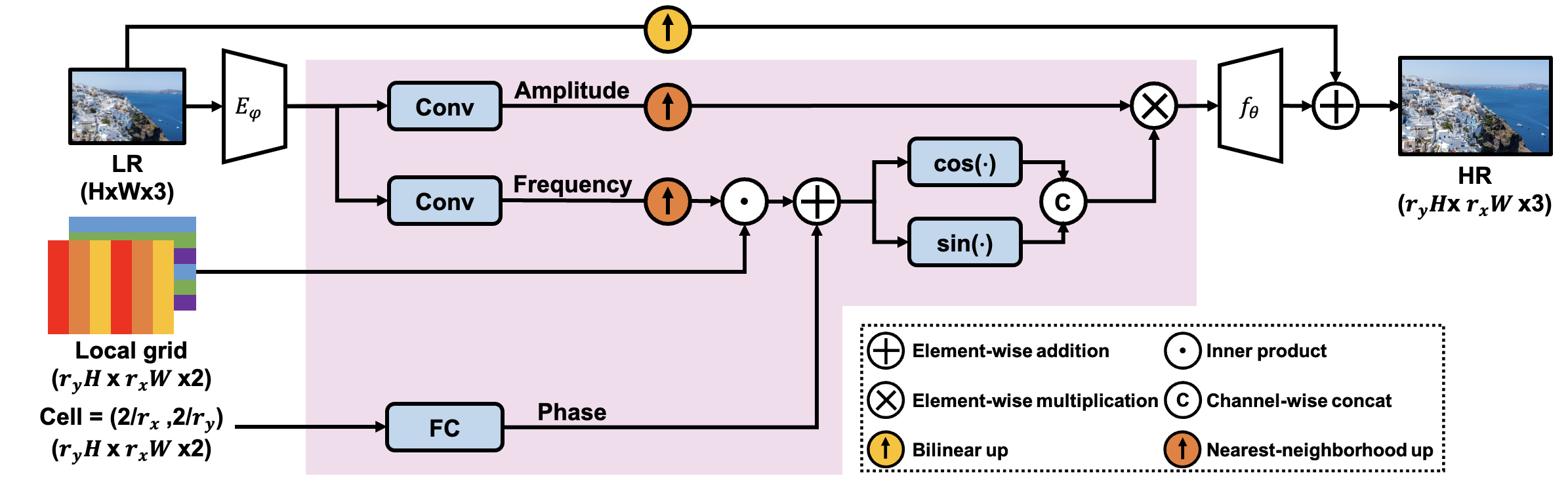}
\caption{\textbf{Arbitrary-scale SR with our proposed Local Texture Estimator (LTE).} LTE-based arbitrary-scale SR architecture consists of an encoder ($E_\varphi$), LTE (a pink shaded region), a decoder ($f_\theta$), and an LR skip connection. Inputs of LTE are as follows: feature map from the encoder, local grid, and cell. LTE transforms input coordinates into the Fourier domain using extracted amplitude, frequency, phase information. We add a bilinear upscaled LR image to the output of a decoder.}
\label{fig:flowchart}
\end{figure*}

{\bf Arbitrary-scale SR} Our work is highly related to SR tasks within an arbitrary-scale factor \cite{hu2019meta, SRWarp, Wang2020Learning, chen2021learning}, which is convenient and efficient for practical benefits. Training and storing models for each specific scale factor \cite{DBLP:journals/corr/ShiCHTABRW16, Lim_2017_CVPR_Workshops, zhang2018residual, zhang2018rcan, DBLP:conf/cvpr/Chen000DLMX0021, liang2021swinir} is unfeasible when considering limited memory resources. Since MetaSR \cite{hu2019meta} first proposed an arbitrary-scale SR method with a single model, various approaches have been explored. Recently, ArbSR \cite{Wang2020Learning} has been proposed as a general plug-in module using conditional convolutions. ArbSR conducts an SR with different scales along horizontal and vertical axes, respectively. More recently, SRWarp \cite{SRWarp} successfully transformed an LR image into any shape in HR representation using a differentiable adaptive warping layer. Most related to ours is the model of \cite{chen2021learning}. Inspired by advancements in implicit neural representation, LIIF replaced sub-pixel convolution with an MLP, taking continuous coordinates and latent variables as inputs. Even though such a method outperformed previous works at large upsampling rates, structural distortion occurs at extreme scales. Instead of concatenating coordinates and latent variables, our LTE-based architecture transforms input coordinates into the Fourier domain using dominant frequencies extracted from latent variables.

\section{Problem Formulation}

In this section, we aim to represent $\mathbf{I^{HR}}\in \mathbb{R}^{r_{\mathbf y}H\times r_{\mathbf x}W\times3}$ from $\mathbf{I^{LR}}\in \mathbb{R}^{H\times W\times3}$ given any fraction number $r_{\mathbf x}, r_{\mathbf y}$. We first review a continuous representation of an RGB image with a local implicit neural representation \cite{Local_Implicit_Grid_CVPR20, chen2021learning}. Even though such approaches showed outstanding performance in representing continuous-domain signals, a standalone MLP fails to capture high-frequency details \cite{DBLP:conf/icml/RahamanBADLHBC19}. To overcome this spectral bias problem, we formulate a Local Texture Estimator (LTE), a dominant frequency estimator for natural images. Unlike prior arts \cite{mildenhall2020nerf, tancik2020fourfeat}, estimated frequencies are data-driven and strongly correlated to image textures. Additionally, we introduce scale-dependent phase estimation and LR skip connection, which aid LTE in learning high-frequency textures.

{\bf Local implicit neural representation} 
In local implicit neural representation \cite{Local_Implicit_Grid_CVPR20, chen2021learning}, a decoding function $f_\theta$ is shared by all images and is parameterized by an MLP with trainable weights $\theta$. A decoder $f_\theta$ maps both latent tensors and local coordinates into RGB values; $f_\theta(\mathbf z,\mathbf x): (\mathcal Z,\mathcal X) \mapsto \mathcal S$. $\mathbf z\in \mathcal Z$ is a latent tensor from an encoder $E_\varphi$, $\mathbf x\in \mathcal X$ is a 2D coordinate in the continuous image domain,  $\mathcal S$ is a space of predicted values from $f_\theta$. For simplicity, we assume that a latent tensor $\mathbf z\in \mathbb{R}^{H\times W\times C}$ has the same width and height as $\mathbf{I^{LR}}$. Then, predicted RGB values ($\mathbf s\in\mathbb{R}^{3}$) at a coordinate $\mathbf{x}\in\mathbb{R}^{2}$ are estimated as
\begin{gather}
    \mathbf s(\mathbf{x},\mathbf{I^{LR}};\Theta)=\sum_{j\in\mathcal{J}}w_j f_\theta(\mathbf{z}_j, \mathbf{x}-\mathbf{x}_j),\label{eq:one}\\
    \mbox{where }\mathbf{z}=  E_\varphi(\mathbf{I^{LR}})\label{eq:two}.
\end{gather}

$\Theta=[\vec \theta;\vec \varphi]$, $\mathcal{J}\in\mathbb{Z}^4$ is a set of indices for four nearest (Euclidean distance) latent codes around $\mathbf{x}$, $w_j$ is the bilinear interpolation weight corresponding to the latent code $j$ (referred to as the local ensemble weight \cite{chen2021learning}, $\sum_j w_j =1$), $\mathbf{z}_j\in \mathbb{R}^{C}$ is the $j-$th nearest latent feature vector from $\mathbf x$, and $\mathbf{x}_j\in \mathbb{R}^{2}$ is the coordinate of the latent code $j$. Given a series of $M$ data points from $N$ images such as $(\mathbf{x}_m,\mathbf I^{\mathbf{ HR}}_n(\mathbf x_m))$, $m=1,\dots,M$ and $n=1,\dots,N$, the learning problem is defined as follows:
\begin{equation}
    \hat{\Theta}=\arg\min_{\Theta} \sum_{m,n}^{M,N}\|\mathbf{I}^{\mathbf{ HR}}_n(\mathbf{x}_m)-\mathbf s(\mathbf{x}_m,\mathbf I^{\mathbf{ LR}}_n;\Theta)\|_1.\label{eq:three}
\end{equation}

In practice, $\mathcal X$ spans $[-H,H]$ and $[-W,W]$ for two dimensions. Note that a step size (\textit{Cell} in \cref{fig:flowchart}) of an output grid ($\mathbb{R}^{r_{\mathbf y}H\times r_{\mathbf x}W}$) and an input grid ($\mathbb{R}^{H\times W}$) is different. In \cite{Local_Implicit_Grid_CVPR20, chen2021learning}, their decoding function ($f_\theta$) predicts continuous representation with a relative coordinate: $\mathbf x -\mathbf x_j$ $(|\mathbf x -\mathbf x_j|\leq\mathbf 1)$ known as \textit{Local grid}. The same coordinate (\textit{Local grid}) with \cite{chen2021learning,Local_Implicit_Grid_CVPR20} is used for our work in order to represent
 a $r_{\mathbf x}\times r_{\mathbf y}$ local area in HR representation.


{\bf Learning dominant frequency component} Recent works have shown that an MLP with ReLUs is biased towards learning low-frequency content \cite{DBLP:conf/icml/RahamanBADLHBC19}. To resolve this spectral bias problem of an implicit neural function, we propose a Local Texture Estimator (LTE), an essential Fourier information estimator for natural images. Inspired by position encoding \cite{mildenhall2020nerf} and Fourier feature mapping \cite{tancik2020fourfeat}, LTE transforms input coordinates into the Fourier domain before passing an MLP. However, unlike \cite{mildenhall2020nerf, tancik2020fourfeat}, estimated Fourier information is data-driven and reflects image textures in 2D Fourier space. The local implicit neural representation in \cref{eq:one} can be modified as follows:
\begin{equation}
    \mathbf s(\mathbf{x},\mathbf{I^{LR}};\Theta,\psi)=\sum_{j\in\mathcal{J}}w_jf_\theta(h_\psi(\mathbf{z}_j,\mathbf{x}-\mathbf{x}_j))
\label{eq:four}
\end{equation}
where $h_\psi(\cdot,\cdot)$ denotes the LTE, which is shift-invariant. LTE ($h_\psi(\cdot,\cdot)$) consists of three elements;(1) an amplitude estimator ($h_a(\cdot):\mathbb{R}^{C}\mapsto \mathbb{R}^{2K}$), (2) a frequency estimator ($h_f(\cdot):\mathbb{R}^{C}\mapsto \mathbb{R}^{K\times2}$), (3) a phase estimator ($h_p(\cdot):\mathbb{R}^{2}\mapsto \mathbb{R}^{K}$). Thus, given a local-grid coordinate $\boldsymbol{\delta}(=\mathbf{x}-\mathbf{x}_j)\in\mathbb{R}^2$, the estimating function $h_\psi(\cdot,\cdot):(\mathbb{R}^C,\mathbb{R}^2)\mapsto \mathbb{R}^{2K}$ is defined as
\begin{gather}
h_\psi(\mathbf{z}_j,\boldsymbol{\delta})=
\mathbf{A}_j\odot\begin{bmatrix}
\cos(\pi \mathbf F_{j}\boldsymbol{\delta})\\
\sin(\pi \mathbf F_{j}\boldsymbol{\delta})
\end{bmatrix},\label{eq:five}\\
\mbox{where }\mathbf{A}_j=h_a(\mathbf z_j),~
\mathbf{F}_j=h_f(\mathbf z_j).
\end{gather}

$\mathbf{A}_j\in\mathbb{R}^{2K}$ is an amplitude vector for a latent code $j$, $\mathbf{F}_j\in\mathbb{R}^{K\times 2}$ denotes a frequency matrix for a latent code $j$, and $\odot$ represents element-wise multiplication. We believe that the amplitude vector and the frequency matrix are extracted from the latent code $j$ so as to represent $\mathbf s(\mathbf x)$ as close as possible to original signals $\mathbf I^{\mathbf{HR}}(\mathbf x)$. In this perspective, we understand that by observing pixels inside a receptive field (RF), LTE with the encoder ($h_\psi \circ E_\varphi$) estimates dominant frequencies and corresponding Fourier coefficients accurately. Here, the size of RF is decided by the encoder ($E_\varphi$). We visually demonstrate estimated frequencies and corresponding Fourier coefficients in \cref{sec:fourier}.


In practice, to enrich the information in outputs of LTE, we apply the unfolding technique to $\mathbf{z}_j$ leading to a concatenation of the $3\times 3$ nearest latent variables in $\hat{\mathbf{z}}_j$ \cite{chen2021learning}. This is implemented with trainable convolutional filters ($h_a(\cdot):\mathbb{R}^{9C}\mapsto \mathbb{R}^{2K},h_f(\cdot):\mathbb{R}^{9C}\mapsto \mathbb{R}^{K\times 2}$).

{\bf Scale-dependent phase estimation} Phase in \cref{eq:seven} contains information about edge locations of features. For SR tasks, the location of edge changes within a small neighborhood in its HR domain when the scale factor changes. To address this issue, we redefine the estimating function as:
\begin{equation}
    h_\psi(\hat{\mathbf z}_j,\boldsymbol{\delta}, \mathbf c)=\mathbf{A}_j\odot\begin{bmatrix}
\cos(\pi(\mathbf{F}_j\boldsymbol{\delta}+h_p(\hat{\mathbf{c}})))\\
\sin(\pi(\mathbf{F}_j\boldsymbol{\delta}+h_p(\hat{\mathbf{c}})))
\end{bmatrix}
\label{eq:seven}
\end{equation}
where $\mathbf{c}$ denotes the cell size. Inspired by the observation that MLPs with ReLUs are incapable of extrapolating unseen non-linear space \cite{DBLP:conf/iclr/XuZLDKJ21}, we use $\hat{\mathbf{c}}=\max(\mathbf{c}, \mathbf{c}_{tr})$, where $\mathbf{c}_{tr}$ denotes the minimum cell size during training.



{\bf LR skip connection} A long skip connection in local implicit representation enriches high-frequency components in residuals and stabilizes convergence \cite{Kim_2016_CVPR}. In the Fourier domain, LTE tends to predict frequencies located near a low-frequency region (DC). To prevent LTE from learning the DC only, we add upscaled LR. Thus, local implicit neural representation with the proposed LTE can be formulated as follows:
\begin{equation}
    \hat{\mathbf s}(\mathbf{x})=\mathbf s(\mathbf{x},\mathbf{I^{LR}};\Theta,\psi)+\mathbf I^{\mathbf{LR}}_{\uparrow}(\mathbf{x})
\label{eq:eight}
\end{equation}

\section{Method}
\subsection{Network Detail}
Our LTE-based arbitrary-scale SR network includes an encoder ($E_\varphi$), the LTE (a pink shaded area in \cref{fig:flowchart}), a decoder ($f_\theta$), and an LR skip connection. This section first describes a backbone structure (including encoder, decoder, and LR skip connection) and our LTE.

\textbf{Backbone} We use EDSR-baseline \cite{Lim_2017_CVPR_Workshops}, RDN \cite{zhang2018residual}, and SwinIR \cite{liang2021swinir} without their upsampling layers as an encoder ($E_\varphi$). Thus, an output of the encoder has the same width and height as an input LR image. We hypothesize that deep SISR networks \cite{Lim_2017_CVPR_Workshops, zhang2018residual, liang2021swinir} aid LTE in estimating crucial Fourier information by extracting features of natural images inside an RF. Our decoder ($f_\theta$), shared by all images as \cite{chen2021learning}, is a 4-layer MLP with a ReLU activation, and its hidden dimensions are 256. Lastly, we add a bilinear upscaled LR image to the decoder output as in \cref{eq:eight}. We expect that such a long skip connection provides DC offsets; thus, LTE is biased toward learning dominant frequencies and corresponding essential Fourier coefficients.

\textbf{LTE} Our LTE contains an amplitude estimator ($h_a$), a frequency estimator ($h_f$), a phase estimator ($h_p$), and sinusoidal activations. An amplitude and a frequency estimator are designed with 3x3 convolutional layers having 256 ($=2K$) output channels, respectively, identical to a fully connected layer when feature maps are unfolded. The phase estimator is a single fully connected layer with hidden dimensions of 128. Note that an amplitude and a frequency estimator take the same feature map while the phase estimator takes cell as an input. Let us assume that a $r_{\mathbf x}\times r_{\mathbf y}$ local region in an HR domain shares amplitude and frequency information extracted from our LTE ($h_\psi$) as in \cref{eq:four}. According to this, we upscale the extracted Fourier information using nearest-neighborhood interpolation. Then, a predicted phase is added to an inner product between the predicted frequency and local grid before passing it through the sinusoidal activation layer, as in \cref{eq:seven}. Finally, we multiply the predicted amplitude and sinusoidal activation output.


\subsection{Training Strategy}
We construct a minibatch with uniformly sampled scales from $\times1-\times4$, dubbed \textit{in-scale}, to teach the nature of bicubic degradation at various scales. Note that we evaluate our LTE for both \textit{in-scale} and \textit{out-of-scale}, which is an unseen scale (specifically $\times6-\times30$), to verify the generalization ability of our network.

Let $r$ ($=r_{\mathbf x}=r_{\mathbf y}$) be a scale factor randomly sampled from $\times1-\times4$ and $H$, $W$ be a height, a width of training patch, respectively. We first crop $rH\times rW$ patches from an HR image. When preparing training pairs, we randomly sample $HW$ pixels from an HR patch for ground truth (GT) and downsample an HR patch by the scale factor $r$ for an LR counterpart. When computing loss during training, we pick $HW$ pixels from interpolation outputs to match the dimensions of prediction with GT.

\begin{figure*}[t]
\footnotesize
\centering
\raisebox{0.2in}{\rotatebox{90}{Urban100($\times8$)}}
\includegraphics[width=1.392in, height=1.044in]{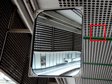}
\includegraphics[width=1.044in]{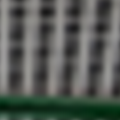}
\includegraphics[width=1.044in]{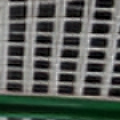}
\includegraphics[width=1.044in]{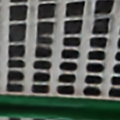}
\includegraphics[width=1.044in]{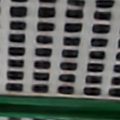}
\includegraphics[width=1.044in]{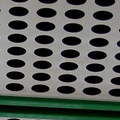}


\raisebox{0.2in}{\rotatebox{90}{DIV2K($\times30$)}}
\stackunder[2pt]{\includegraphics[width=1.392in, height=1.044in]{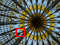}}{LR Image}
\stackunder[2pt]{\includegraphics[width=1.044in]{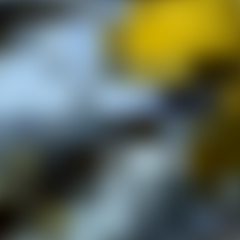}}{Bicubic}
\stackunder[2pt]{\includegraphics[width=1.044in]{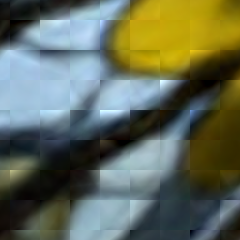}}{MetaSR \cite{hu2019meta, chen2021learning}}
\stackunder[2pt]{\includegraphics[width=1.044in]{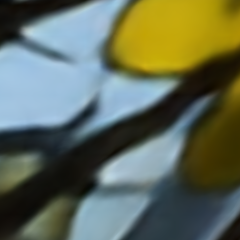}}{LIIF \cite{chen2021learning}}
\stackunder[2pt]{\includegraphics[width=1.044in]{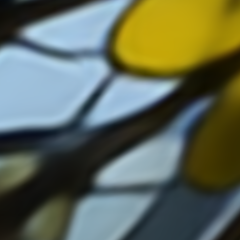}}{\textbf{LTE} (ours)}
\stackunder[2pt]{\includegraphics[width=1.044in]{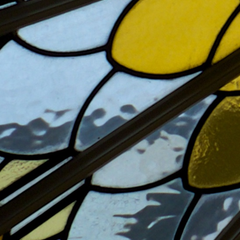}}{GT}



\vspace*{-6pt}
\caption{Qualitative comparison to other \underline{\textbf{arbitrary-scale SR}}. RDN \cite{zhang2018residual} is used as an encoder for all methods.}
\label{fig:Qual_dec_ext}
\end{figure*}

\begin{figure*}[t]
\footnotesize
\centering

\includegraphics[scale=1.9]{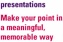}
\includegraphics[scale=1.9]{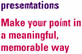}
\includegraphics[scale=1.9]{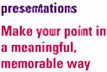}
\includegraphics[scale=1.9]{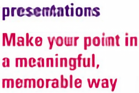}
\includegraphics[scale=1.9]{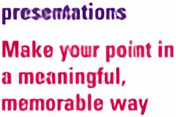}
\includegraphics[scale=1.9]{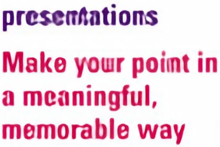}
\includegraphics[scale=1.9]{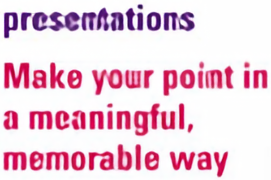}

\stackunder[2pt]{\includegraphics[scale=1.9]{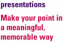}}{\textbf{Input}}
\stackunder[2pt]{\includegraphics[scale=1.9]{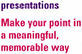}}{$\times$\textbf{1.3}}
\stackunder[2pt]{\includegraphics[scale=1.9]{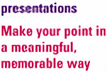}}{$\times$\textbf{1.7}}
\stackunder[2pt]{\includegraphics[scale=1.9]{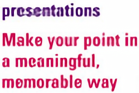}}{$\times$\textbf{2.2}}
\stackunder[2pt]{\includegraphics[scale=1.9]{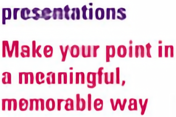}}{$\times$\textbf{2.8}}
\stackunder[2pt]{\includegraphics[scale=1.9]{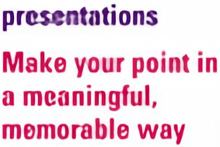}}{$\times$\textbf{3.5}}
\stackunder[2pt]{\includegraphics[scale=1.9]{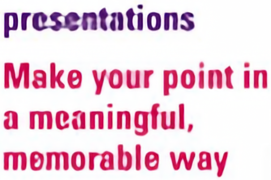}}{$\times$\textbf{4.3}}

\vspace*{-6pt}
\caption{Qualitative comparison with \underline{\textbf{non-integer scale factors}} by LIIF \cite{chen2021learning} (top) and our LTE (bottom). RDN \cite{zhang2018residual} is used as an encoder.}
\label{fig:Qual_dec_arb}
\end{figure*}

\begin{figure}[t]
\footnotesize
\centering
\stackunder[2pt]{\includegraphics[width=0.8in]{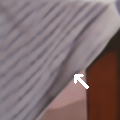}}{EDSR \cite{Lim_2017_CVPR_Workshops}}
\stackunder[2pt]{\includegraphics[width=0.8in]{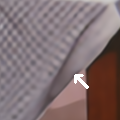}}{RDN \cite{zhang2018residual}}
\stackunder[2pt]{\includegraphics[width=0.8in]{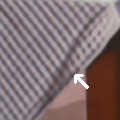}}{SwinIR \cite{liang2021swinir}}
\stackunder[2pt]{\includegraphics[width=0.8in]{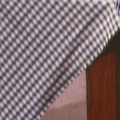}}{GT}%

\vspace*{-6pt}
\caption{Qualitative comparison between various encoders with our LTE for \underline{\textbf{$\times$6 SR}}.}
\label{fig:enc}
\end{figure}

\section{Experiment}

\subsection{Training}
\textbf{Dataset} We use a DIV2K dataset \cite{8014884} of an NTIRE 2017 Challenge \cite{8014883} for network training. For evaluation, we report peak signal-to-noise ratio (PSNR) results on the DIV2K validation set \cite{8014884}, Set5 \cite{bevilacqua2012low}, Set14 \cite{zeyde2010single}, B100 \cite{martin2001database}, and Urban100 \cite{7299156}.

\textbf{Implementation detail} We follow a prior implementation \cite{Lim_2017_CVPR_Workshops} and use $48\times48$ patches for inputs of our network. For arbitrary-scale down-sampling during training time, we follow \cite{chen2021learning} and use bicubic resizing in Pytorch \cite{paszke2019pytorch}. We use an L1 loss \cite{Lim_2017_CVPR_Workshops} and an Adam \cite{DBLP:journals/corr/KingmaB14} method for optimization. When we train LTE with CNN-based encoders, such as EDSR-baseline \cite{Lim_2017_CVPR_Workshops} or RDN \cite{zhang2018residual}, networks are trained for 1000 epochs with batch size 16. The learning rate is initialized as 1e-4 and decayed by factor 0.5 at [200, 400, 600, 800]. For a transformer-based encoder, specifically SwinIR \cite{liang2021swinir}, a model is trained for 1000 epochs with batch size 32. The learning rate is initialized as 2e-4 and decayed by factor 0.5 at [500, 800, 900, 950].

\begin{table*}[t]
\centering
\setlength{\tabcolsep}{1.2pt}
\scriptsize{
\begin{tabular}{c
|>{\centering\arraybackslash}p{0.67cm}>{\centering\arraybackslash}p{0.67cm}>{\centering\arraybackslash}p{0.67cm}
|>{\centering\arraybackslash}p{0.67cm}>{\centering\arraybackslash}p{0.67cm}
>{\centering\arraybackslash}p{0.67cm}>{\centering\arraybackslash}p{0.67cm}>{\centering\arraybackslash}p{0.67cm}}
\multirow{2}{*}{Method} & \multicolumn{3}{c|}{In-scale} & \multicolumn{5}{c}{Out-of-scale} \\
 & $\times2$ & $\times3$ & $\times4$ & $\times6$ & $\times12$ & $\times18$ & $\times24$ & $\times30$\\
\hline\hline
Bicubic \cite{Lim_2017_CVPR_Workshops} & 31.01 & 28.22 & 26.66 & 24.82  & 22.27 & 21.00 & 20.19 & 19.59 \\
EDSR-baseline \cite{Lim_2017_CVPR_Workshops} & 34.55 & 30.90 & 28.94 & - & - & - & - & - \\
EDSR-baseline-MetaSR \cite{hu2019meta, chen2021learning} & 34.64 & 30.93 & 28.92 & 26.61 & 23.55 & 22.03 & 21.06 & 20.37 \\
EDSR-baseline-LIIF \cite{chen2021learning} & \textcolor{blue}{34.67} & \textcolor{blue}{30.96} & \textcolor{blue}{29.00} & \textcolor{blue}{26.75} & \textcolor{blue}{23.71} & \textcolor{blue}{22.17} & \textcolor{blue}{21.18} & \textcolor{blue}{20.48} \\
EDSR-baseline-LTE (ours) & \textcolor{red}{34.72} & \textcolor{red}{31.02} & \textcolor{red}{29.04} & \textcolor{red}{26.81} & \textcolor{red}{23.78} & \textcolor{red}{22.23} & \textcolor{red}{21.24} & \textcolor{red}{20.53} \\
\hline
RDN-MetaSR \cite{hu2019meta, chen2021learning} & \textcolor{blue}{35.00} & \textcolor{blue}{31.27} & 29.25 & 26.88 & 23.73 & 22.18 & 21.17 & 20.47 \\
RDN-LIIF \cite{chen2021learning} & 34.99 & 31.26 & \textcolor{blue}{29.27} & \textcolor{blue}{26.99} & \textcolor{blue}{23.89} & \textcolor{blue}{22.34} & \textcolor{blue}{21.31} & \textcolor{blue}{20.59} \\
RDN-LTE (ours) & \textcolor{red}{35.04} & \textcolor{red}{31.32} & \textcolor{red}{29.33} & \textcolor{red}{27.04} & \textcolor{red}{23.95} & \textcolor{red}{22.40} & \textcolor{red}{21.36} & \textcolor{red}{20.64} \\
\hline
SwinIR-MetaSR$^{\dag}$ \cite{hu2019meta, chen2021learning} & 35.15 & 31.40 & 29.33 & 26.94 & 23.80 & 22.26 & 21.26 & 20.54 \\
SwinIR-LIIF$^{\dag}$ \cite{chen2021learning} & \textcolor{blue}{35.17} & \textcolor{blue}{31.46} & \textcolor{blue}{29.46} & \textcolor{blue}{27.15} & \textcolor{blue}{24.02} & \textcolor{blue}{22.43} & \textcolor{blue}{21.40} & \textcolor{blue}{20.67} \\
Swinir-LTE (ours) & \textcolor{red}{35.24} & \textcolor{red}{31.50} & \textcolor{red}{29.51} & \textcolor{red}{27.20} & \textcolor{red}{24.09} & \textcolor{red}{22.50} & \textcolor{red}{21.47} & \textcolor{red}{20.73} \\
\end{tabular}
}
\vspace*{-6pt}
\caption{Quantitative comparison with state-of-the-art methods for \underline{\textbf{arbitrary-scale SR}} on DIV2K validation set (PSNR (dB)). \textcolor{red}{Red} and \textcolor{blue}{blue} colors indicate the best and the second-best performance, respectively. $\dag$ indicates our implementation.}
\label{tab:Quan_DIV2K}
\end{table*}

\begin{table*}[ht]
\centering
\setlength{\tabcolsep}{1.2pt}
\scriptsize{
\begin{tabular}{c
|>{\centering\arraybackslash}p{0.67cm}>{\centering\arraybackslash}p{0.67cm}>{\centering\arraybackslash}p{0.67cm}
|>{\centering\arraybackslash}p{0.67cm}>{\centering\arraybackslash}p{0.67cm}
|>{\centering\arraybackslash}p{0.67cm}>{\centering\arraybackslash}p{0.67cm}>{\centering\arraybackslash}p{0.67cm}
|>{\centering\arraybackslash}p{0.67cm}>{\centering\arraybackslash}p{0.67cm}
|>{\centering\arraybackslash}p{0.67cm}>{\centering\arraybackslash}p{0.67cm}>{\centering\arraybackslash}p{0.67cm}
|>{\centering\arraybackslash}p{0.67cm}>{\centering\arraybackslash}p{0.67cm}
|>{\centering\arraybackslash}p{0.67cm}>{\centering\arraybackslash}p{0.67cm}>{\centering\arraybackslash}p{0.67cm}
|>{\centering\arraybackslash}p{0.67cm}>{\centering\arraybackslash}p{0.67cm}}
\multirow{3}{*}{Method} & \multicolumn{5}{c|}{Set5} & \multicolumn{5}{c|}{Set14}
& \multicolumn{5}{c|}{B100} & \multicolumn{5}{c}{Urban100} \\
\cline{2-21}
& \multicolumn{3}{c|}{In-scale} & \multicolumn{2}{c|}{Out-of-scale}
& \multicolumn{3}{c|}{In-scale} & \multicolumn{2}{c|}{Out-of-scale}
& \multicolumn{3}{c|}{In-scale} & \multicolumn{2}{c|}{Out-of-scale}
& \multicolumn{3}{c|}{In-scale} & \multicolumn{2}{c}{Out-of-scale} \\
& $\times2$ & $\times3$ & $\times4$ & $\times6$ & $\times8$
& $\times2$ & $\times3$ & $\times4$ & $\times6$ & $\times8$
& $\times2$ & $\times3$ & $\times4$ & $\times6$ & $\times8$
& $\times2$ & $\times3$ & $\times4$ & $\times6$ & $\times8$\\
\hline\hline
RDN \cite{zhang2018residual} & \textcolor{red}{38.24} & \textcolor{blue}{34.71} & 32.47 & - & -
& \textcolor{blue}{34.01} & \textcolor{blue}{30.57} & \textcolor{blue}{28.81} & - & -
& \textcolor{blue}{32.34} & \textcolor{blue}{29.26} & 27.72 & - & -
& 32.89 & 28.80 & 26.61 & - & - \\
RDN-MetaSR \cite{hu2019meta, chen2021learning} & 38.22 & 34.63 & 32.38 & 29.04 & 26.96
& 33.98 & 30.54 & 28.78 & 26.51 & 24.97
& 32.33 & \textcolor{blue}{29.26} & 27.71 & 25.90 & 24.83
& \textcolor{blue}{32.92} & \textcolor{blue}{28.82} & 26.55 & 23.99 & 22.59 \\
RDN-LIIF \cite{chen2021learning} & 38.17 & 34.68 & \textcolor{blue}{32.50} & \textcolor{blue}{29.15} & \textcolor{blue}{27.14}
& 33.97 & 30.53 & 28.80 & \textcolor{blue}{26.64} & \textcolor{blue}{25.15}
& 32.32 & \textcolor{blue}{29.26} & \textcolor{blue}{27.74} & \textcolor{blue}{25.98} & \textcolor{blue}{24.91}
& 32.87 & \textcolor{blue}{28.82} & \textcolor{blue}{26.68} & \textcolor{blue}{24.20} & \textcolor{blue}{22.79} \\
RDN-LTE (ours) & \textcolor{blue}{38.23} & \textcolor{red}{34.72} & \textcolor{red}{32.61} & \textcolor{red}{29.32} & \textcolor{red}{27.26}
& \textcolor{red}{34.09} & \textcolor{red}{30.58} & \textcolor{red}{28.88} & \textcolor{red}{26.71} & \textcolor{red}{25.16}
& \textcolor{red}{32.36} & \textcolor{red}{29.30} & \textcolor{red}{27.77} & \textcolor{red}{26.01} & \textcolor{red}{24.95}
& \textcolor{red}{33.04} & \textcolor{red}{28.97} & \textcolor{red}{26.81} & \textcolor{red}{24.28} & \textcolor{red}{22.88} \\
\hline
SwinIR \cite{liang2021swinir} & \textcolor{red}{38.35} & \textcolor{blue}{34.89} & \textcolor{black}{32.72} & - & -
& \textcolor{blue}{34.14} & \textcolor{blue}{30.77} & \textcolor{black}{28.94} & - & - 
& \textcolor{blue}{32.44} & \textcolor{blue}{29.37} & \textcolor{black}{27.83} & - & -
& \textcolor{blue}{33.40} & \textcolor{black}{29.29} & \textcolor{black}{27.07} & - & - \\
SwinIR-MetaSR$^{\dag}$ \cite{hu2019meta, chen2021learning}
& \textcolor{black}{38.26} & \textcolor{black}{34.77} & \textcolor{black}{32.47}
& \textcolor{black}{29.09} & \textcolor{black}{27.02}
& \textcolor{blue}{34.14} & \textcolor{black}{30.66} & \textcolor{black}{28.85}
& \textcolor{black}{26.58} & \textcolor{black}{25.09}
& \textcolor{black}{32.39} & \textcolor{black}{29.31} & \textcolor{black}{27.75}
& \textcolor{black}{25.94} & \textcolor{black}{24.87}
& \textcolor{black}{33.29} & \textcolor{black}{29.12} & \textcolor{black}{26.76}
& \textcolor{black}{24.16} & \textcolor{black}{22.75} \\
SwinIR-LIIF$^{\dag}$ \cite{chen2021learning}
& \textcolor{black}{38.28} & \textcolor{black}{34.87} & \textcolor{blue}{32.73}
& \textcolor{blue}{29.46} & \textcolor{red}{27.36}
& \textcolor{blue}{34.14} & \textcolor{black}{30.75} & \textcolor{blue}{28.98}
& \textcolor{blue}{26.82} & \textcolor{blue}{25.34}
& \textcolor{black}{32.39} & \textcolor{black}{29.34} & \textcolor{blue}{27.84}
& \textcolor{blue}{26.07} & \textcolor{blue}{25.01}
& \textcolor{black}{33.36} & \textcolor{blue}{29.33} & \textcolor{blue}{27.15}
& \textcolor{blue}{24.59} & \textcolor{blue}{23.14} \\
SwinIR-LTE (ours) & \textcolor{blue}{38.33} & \textcolor{blue}{34.89} & \textcolor{red}{32.81} & \textcolor{red}{29.50} & \textcolor{blue}{27.35}
& \textcolor{red}{34.25} & \textcolor{red}{30.80} & \textcolor{red}{29.06} & \textcolor{red}{26.86} & \textcolor{red}{25.42}
& \textcolor{blue}{32.44} & \textcolor{red}{29.39} & \textcolor{red}{27.86} & \textcolor{red}{26.09} & \textcolor{red}{25.03}
& \textcolor{red}{33.50} & \textcolor{red}{29.41} & \textcolor{red}{27.24} & \textcolor{red}{24.62} & \textcolor{red}{23.17} \\
\end{tabular}
}
\vspace*{-6pt}
\caption{Quantitative comparison with state-of-the-art methods for \underline{\textbf{arbitrary-scale SR}} on benchmark datasets (PSNR (dB)). \textcolor{red}{Red} and \textcolor{blue}{blue} colors indicate the best and the second-best performance, respectively. $\dag$ indicates our implementation.}
\label{tab:Quan_Bench}
\vspace{-10pt}
\end{table*}

\begin{figure}[t]
\footnotesize
\centering

\stackunder[2pt]{\includegraphics[width=1.06in]{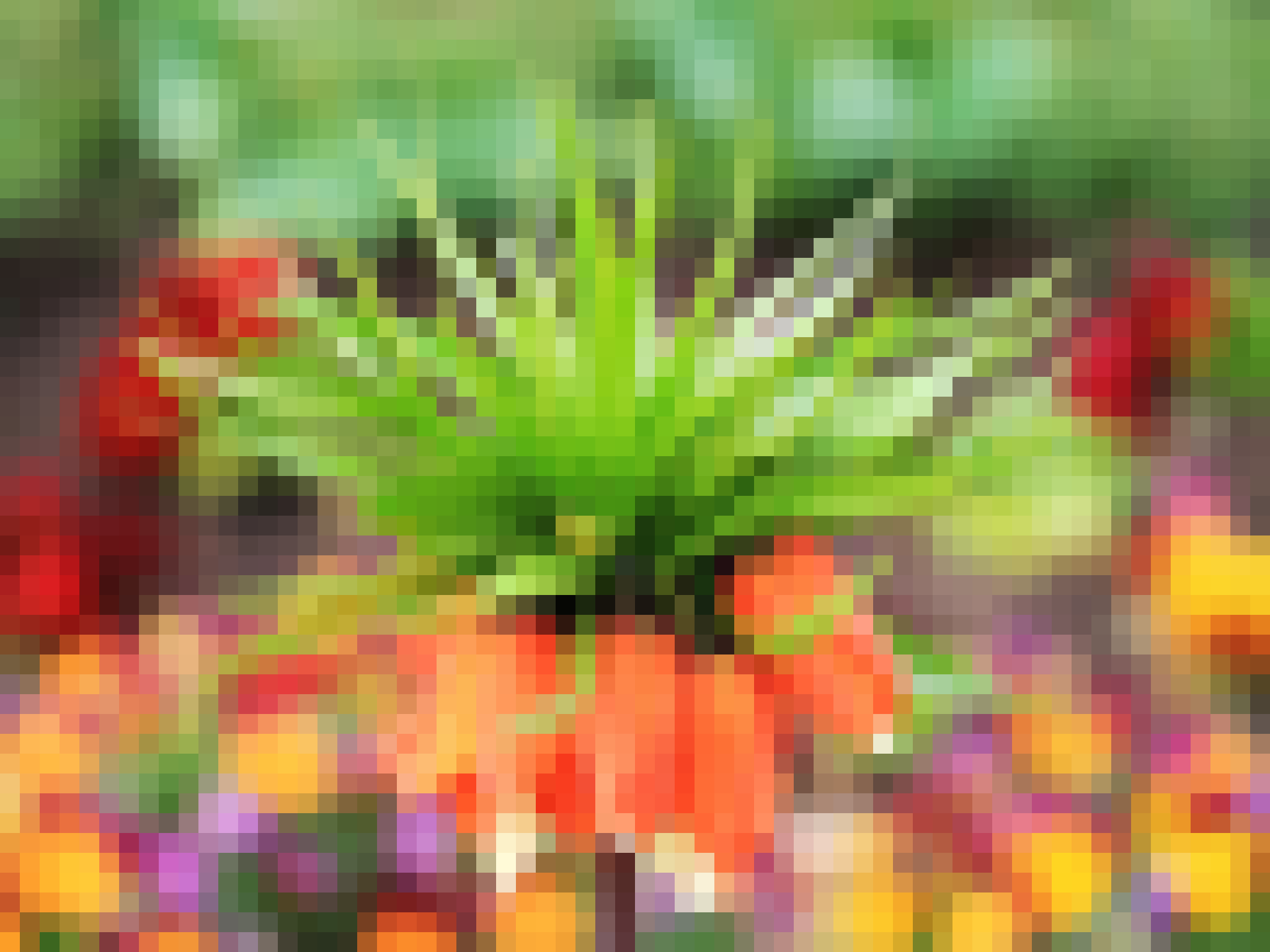}}{Input}
\stackunder[2pt]{\includegraphics[width=1.06in]{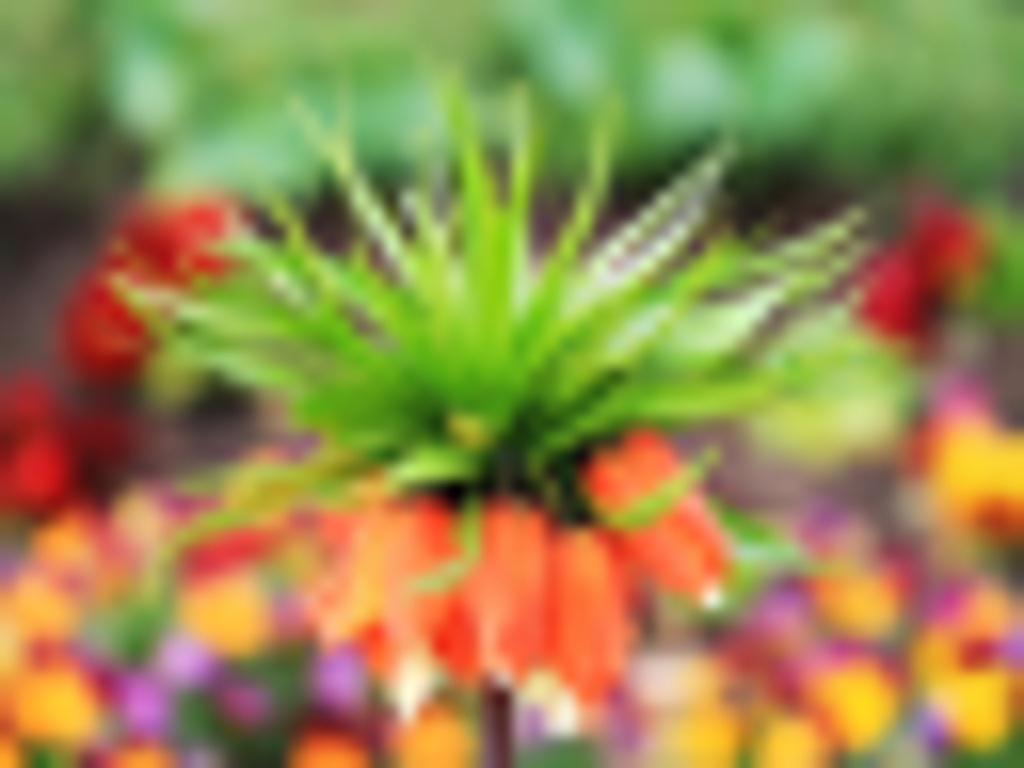}}{Bicubic}
\stackunder[2pt]{\includegraphics[width=1.06in]{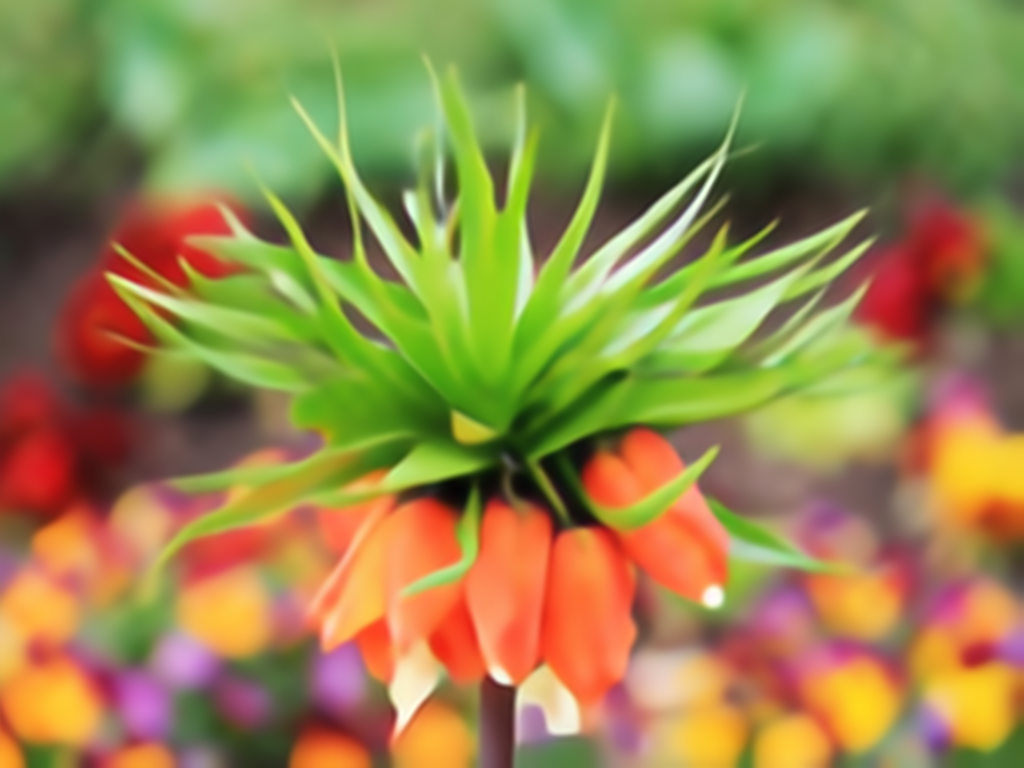}}{SwinIR-LTE}

\vspace*{-6pt}
\caption{Visual demonstration of SwinIR-LTE for \underline{\textbf{$\times$32 SR}}.}
\label{fig:demo_ext}
\end{figure}

\subsection{Evaluation}
\label{sec:eval}
\textbf{Quantitative result} \cref{tab:Quan_DIV2K} demonstrates a quantitative comparison between our LTE and existing arbitrary-scale SR methods, MetaSR \cite{hu2019meta}, LIIF \cite{chen2021learning}, on the DIV2K validation set. The top, middle, and bottom rows show results when EDSR-baseline \cite{Lim_2017_CVPR_Workshops}, RDN \cite{zhang2018residual}, and SwinIR \cite{liang2021swinir} are used as encoders. We notice that, regardless of a choice of an encoder, LTE achieves the best performance for all the scale factors, which indicates the effectiveness of the local texture.

In \cref{tab:Quan_Bench}, we compare our LTE and RDN \cite{zhang2018residual}, SwinIR \cite{liang2021swinir}, MetaSR \cite{hu2019meta}, LIIF \cite{chen2021learning} on benchmark datasets. Note that RDN and SwinIR \cite{liang2021swinir} are trained with a specific scale; thus, it has significant benefits for \textit{in-scale} \cite{chen2021learning}. However, including RDN and SwinIR, our LTE shows remarkable performance compared to other methods. The maximum PSNR gain is 0.15dB on Urban100 for $\times3$ within RDN.

\textbf{Qualitative result} Qualitative comparisons to other arbitrary-scale SR methods are provided in \cref{fig:Qual_dec_ext}. For a fair comparison, MetaSR \cite{hu2019meta}, LIIF \cite{chen2021learning}, and our LTE are trained with RDN \cite{zhang2018residual}. Note that MetaSR \cite{hu2019meta} follows \cite{chen2021learning}'s implementation to reconstruct an HR image for large-scale factors ($>\times4$). We see that MetaSR suffers from blocky artifacts, and LIIF shows structural distortion. In contrast, our LTE captures high-frequency details without any discontinuities.

\cref{fig:Qual_dec_arb} compares LIIF \cite{chen2021learning} and our LTE for text images with non-integer scale factors. We observe that our LTE is capable of restoring more clear edges of printed texts for all scale factors (particularly, `n', `t', `s' in the first row and `u', `i', `n' in the second row).

\begin{figure}[t]
\vspace{-6pt}
\footnotesize
\centering
\stackunder[2pt]{\includegraphics[width=0.972in, height=0.729in]{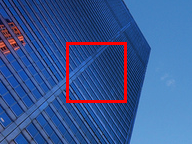}}{GT}
\stackunder[2pt]{\includegraphics[height=0.729in]{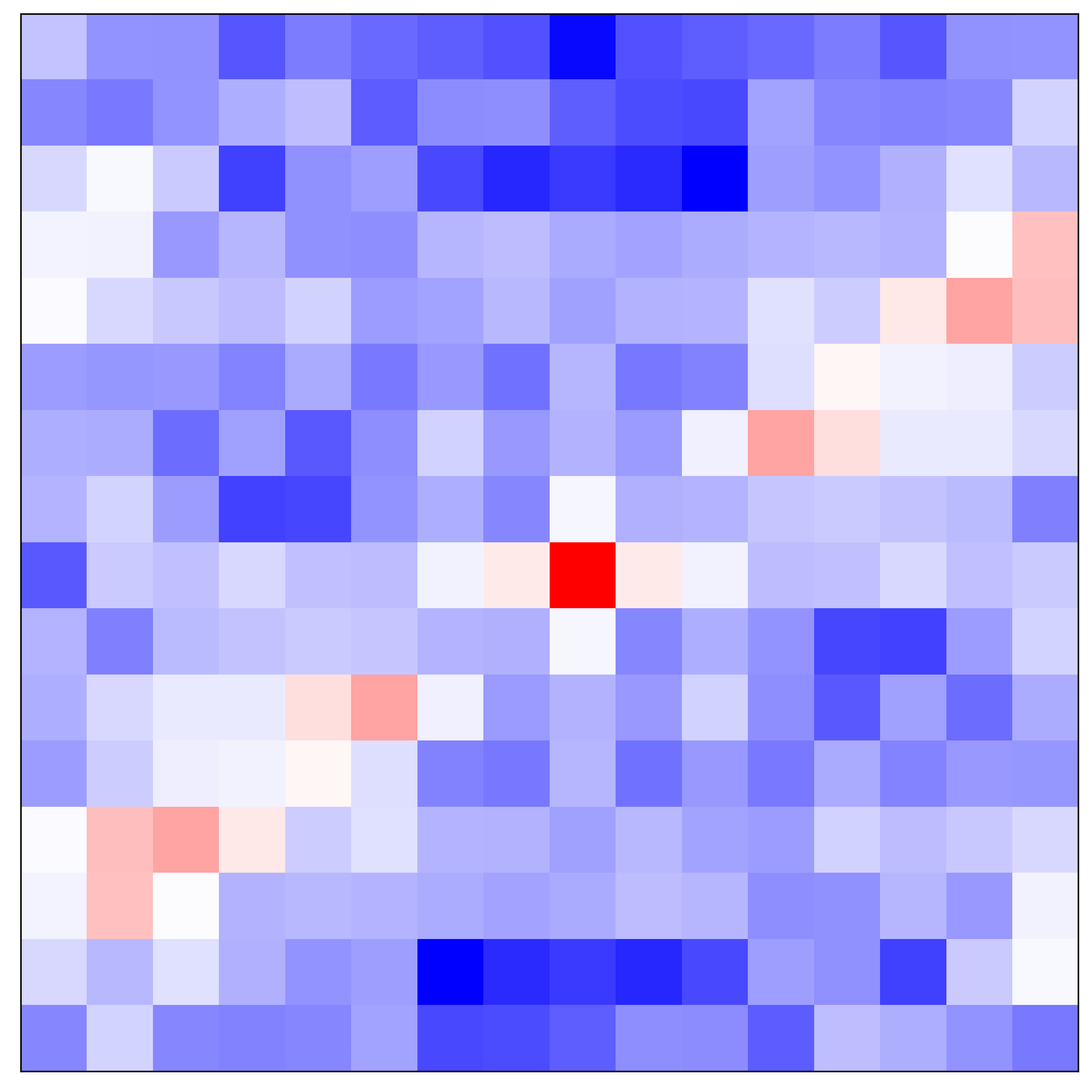}}{GT spectrum}
\stackunder[2pt]{\includegraphics[height=0.729in]{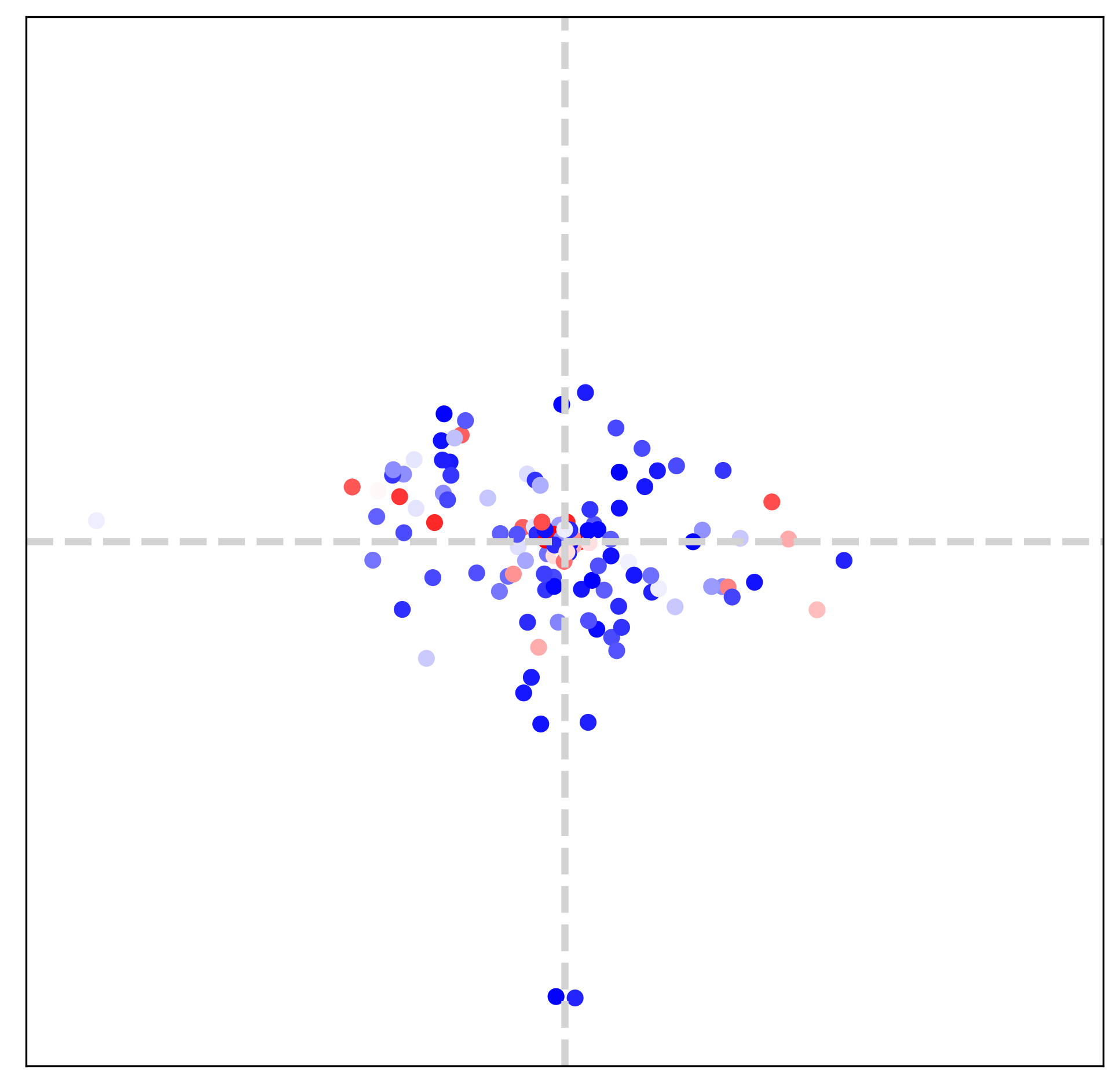}}{EDSR \cite{Lim_2017_CVPR_Workshops}}
\stackunder[2pt]{\includegraphics[height=0.729in]{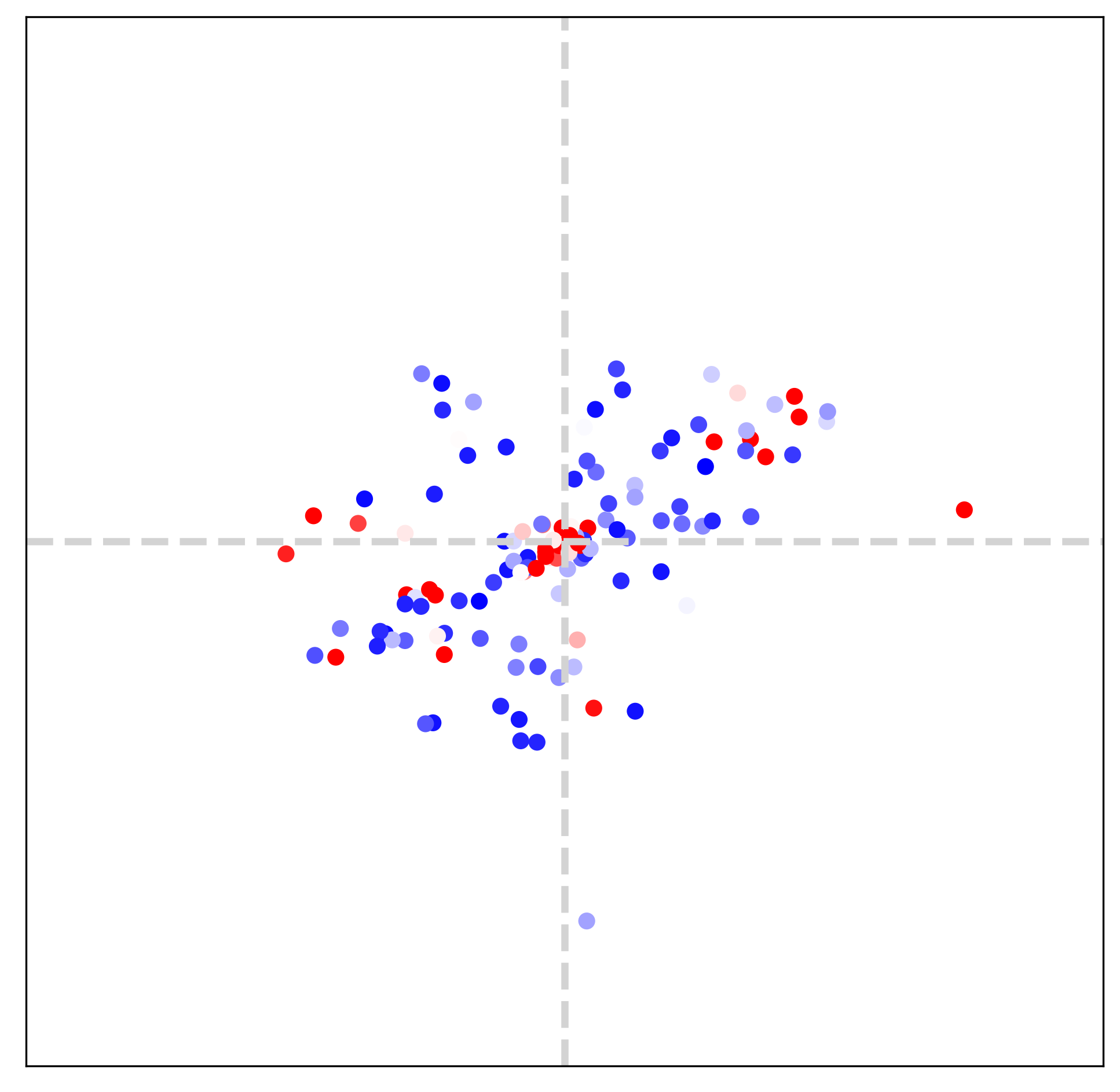}}{SwinIR \cite{liang2021swinir}}

\vspace*{-6pt}
\caption{Estimated Fourier space with various encoders.}
\label{fig:four:enc}
\vspace{-15pt}
\end{figure}

In \cref{fig:enc}, we show a qualitative comparison for $\times6$ SR. We remark that SwinIR \cite{liang2021swinir} followed by LTE reconstructs the most visually pleasing image, faithful to the GT. It implies that LTE precisely extracts dominant frequencies and corresponding essential Fourier coefficients when jointly trained with a robust encoder. An empirical explanation using Fourier analysis is provided in \cref{fig:four:enc} and \cref{sec:fourier}.

As shown in \cref{fig:demo_ext}, we visually demonstrate our LTE at an extremely large scale factor, specifically $\times32$. For the demonstration, we trained our LTE with SwinIR \cite{liang2021swinir}, and the width of an input image is 64px. We note that our LTE interpolates images with more sharp and natural edges compared to the bicubic method.

\subsection{Ablation Study}
\label{sec:abl}
In this section, we demonstrate the effect of each component in LTE. Our LTE consists of an amplitude estimator, a frequency estimator, a phase estimator, and an LR skip connection. To support the significance of each component, we retrain the following models with EDSR-baseline \cite{Lim_2017_CVPR_Workshops}. (-A): LTE without an amplitude estimator. (-F): LTE with a frequency estimator that estimates only 128 frequencies (not 256). (-P): LTE without a phase estimator. (-L): LTE without an LR skip connection.

\cref{fig:qual_abl} and \cref{tab:Quan_abl} show the contributions of each LTE component on visual quality and performance. To verify the importance of each estimated frequency, we compare LTE to LTE (-F). We find that an amplitude estimator emphasizes dominant frequencies compared between LTE and LTE (-A). By comparing LTE and LTE (-P), disregarding a phase difference causes a significant performance drop. We see that an LR skip connection consistently enhances the quality of LTE when comparing LTE with LTE (-L).

\begin{figure}[t]
\footnotesize
\centering
\includegraphics[width=1.06in, height=0.795in]{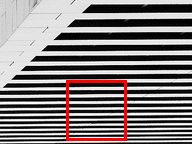}
\includegraphics[width=1.06in, height=0.795in]{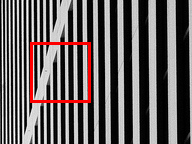}
\includegraphics[width=1.06in, height=0.795in]{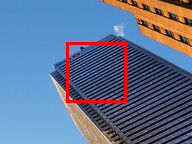}
\includegraphics[width=1.06in]{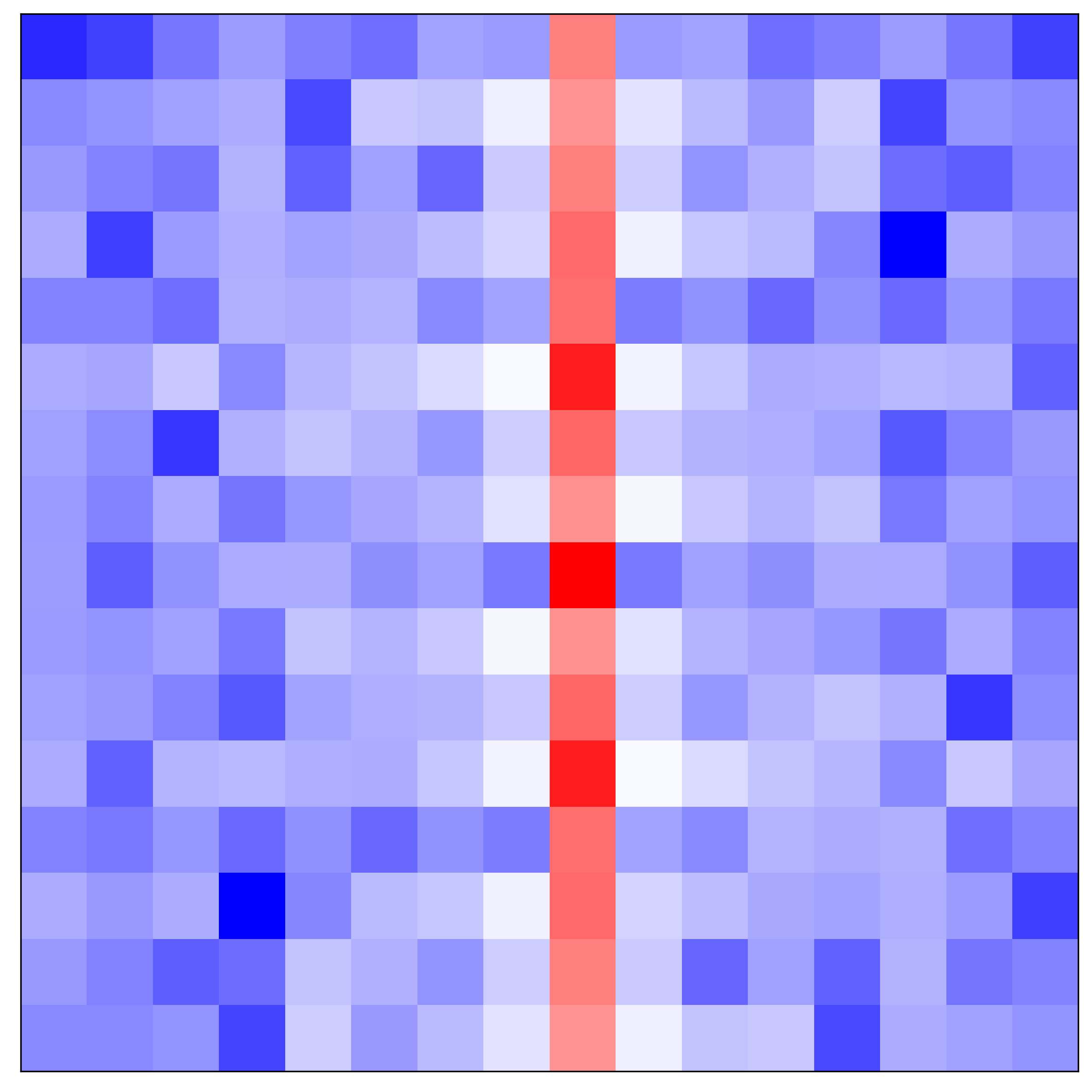}
\includegraphics[width=1.06in]{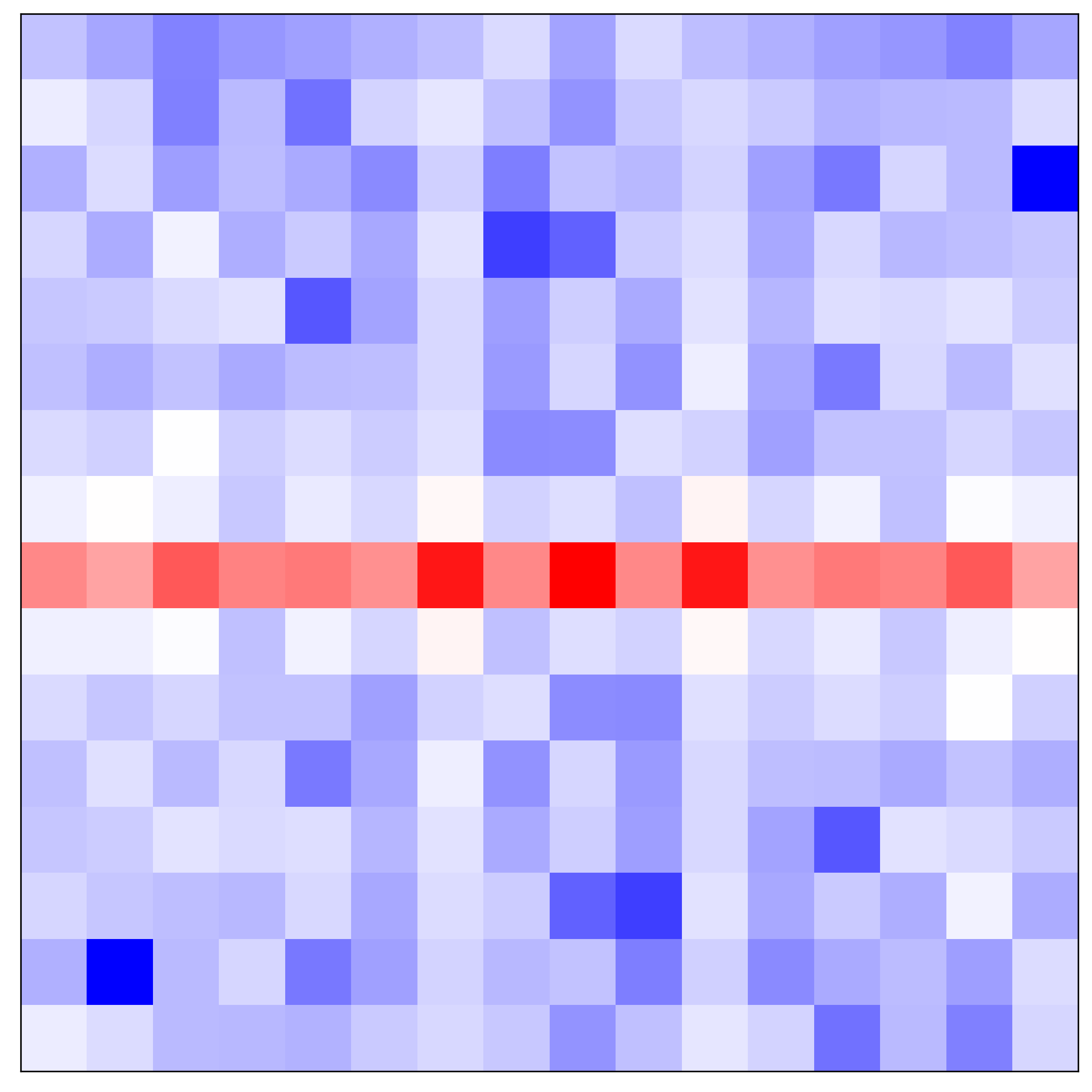}
\includegraphics[width=1.06in]{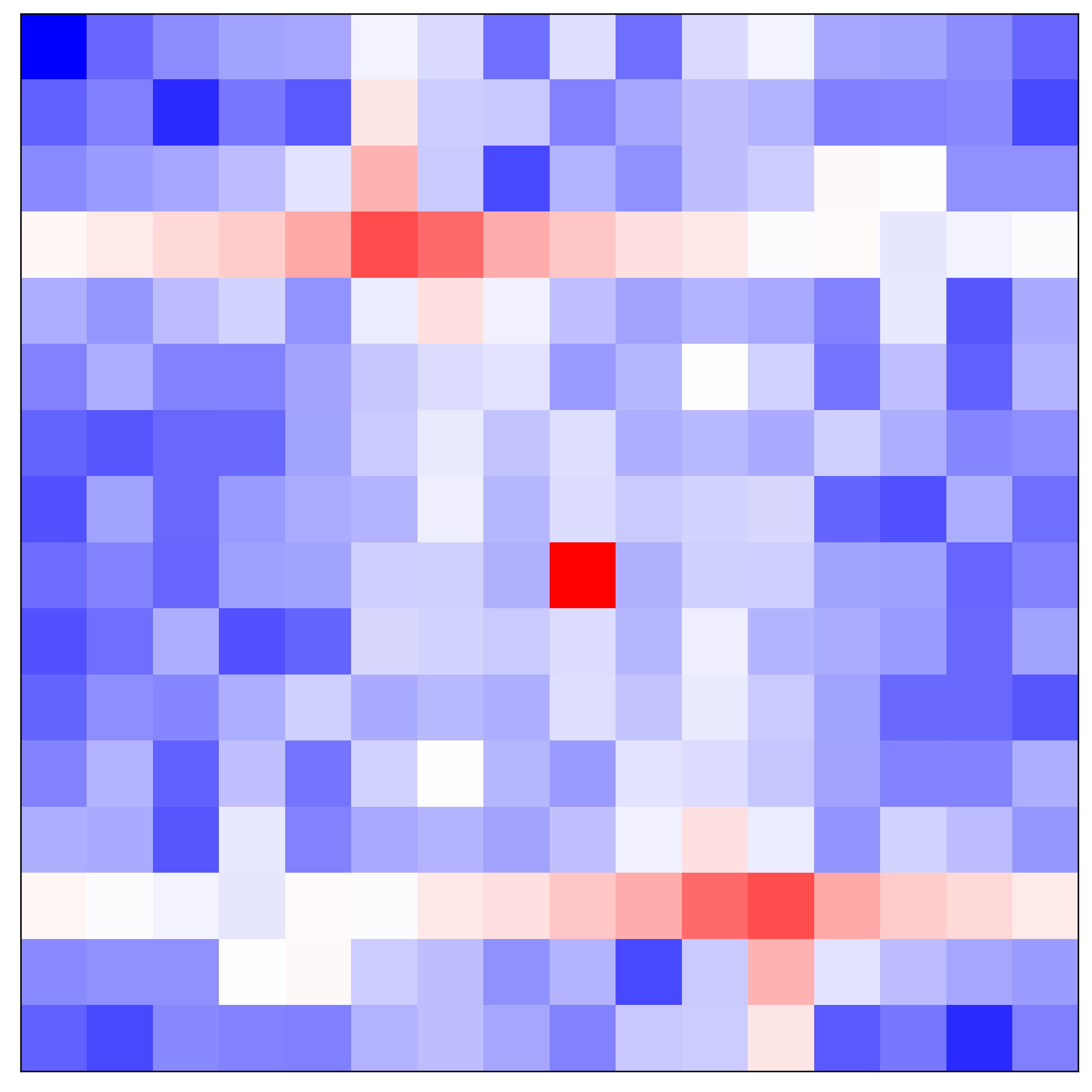}
\stackunder[2pt]{\includegraphics[width=1.06in]{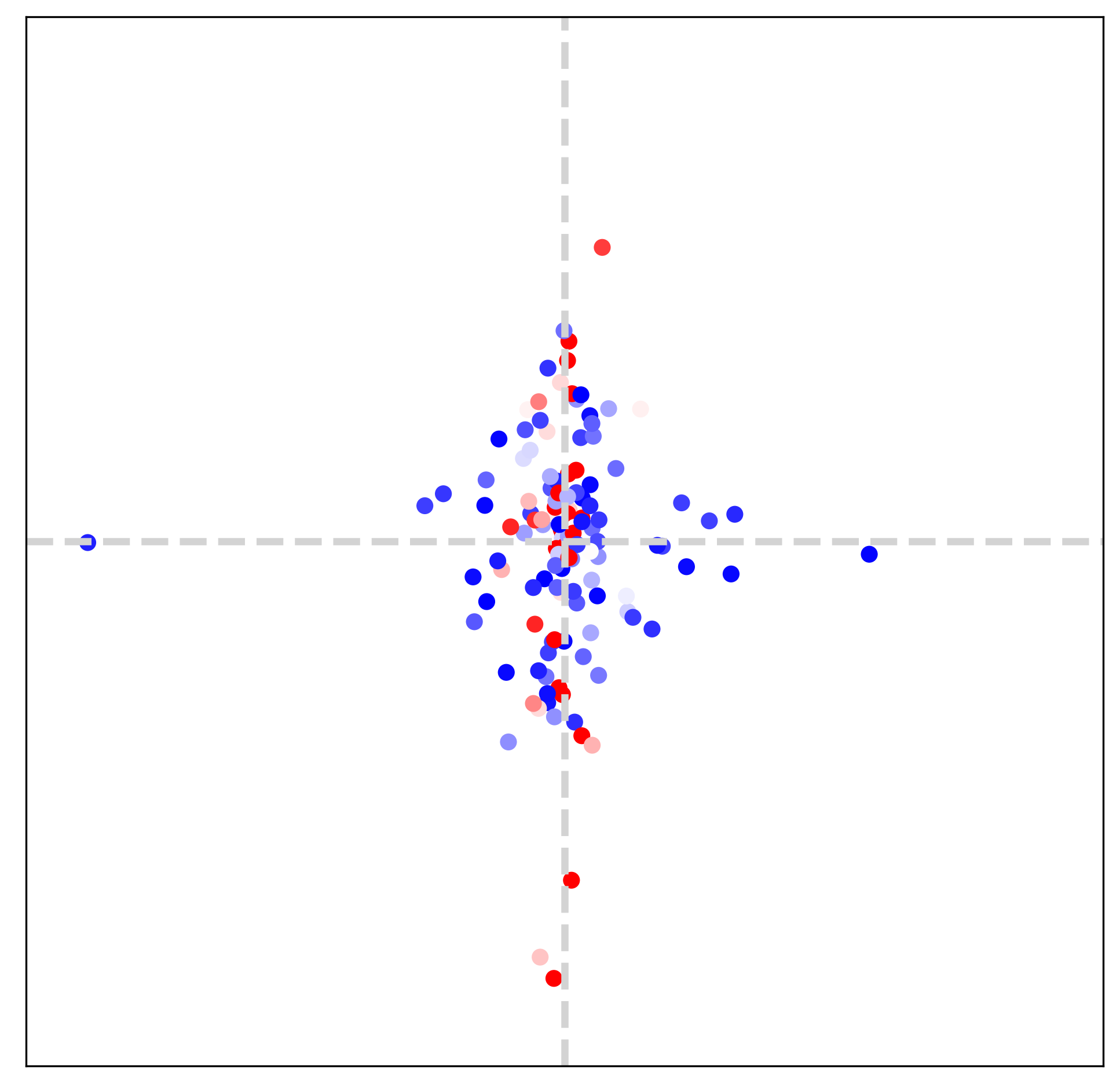}}{Horizontal}
\stackunder[2pt]{\includegraphics[width=1.06in]{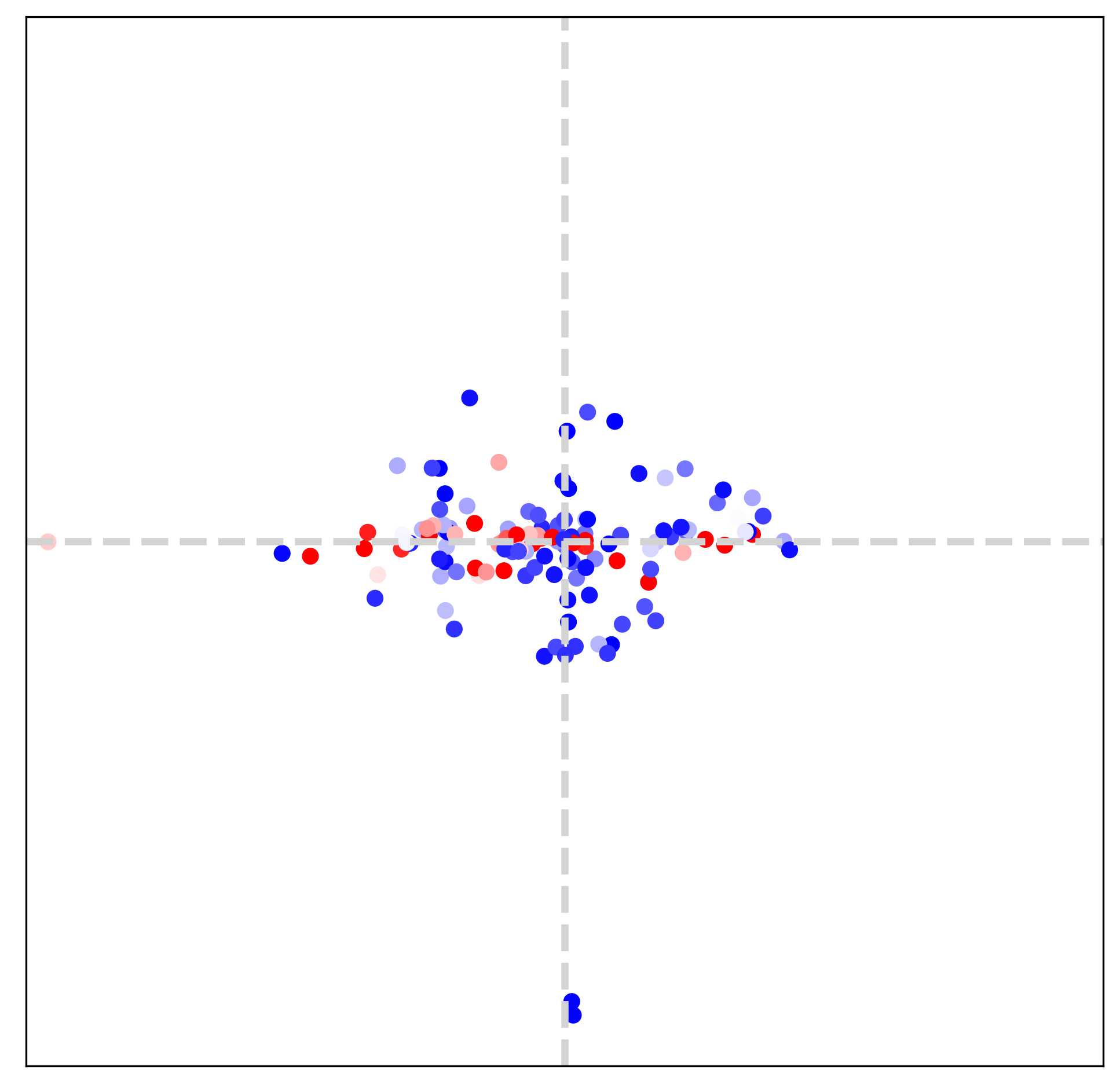}}{Vertical}
\stackunder[2pt]{\includegraphics[width=1.06in]{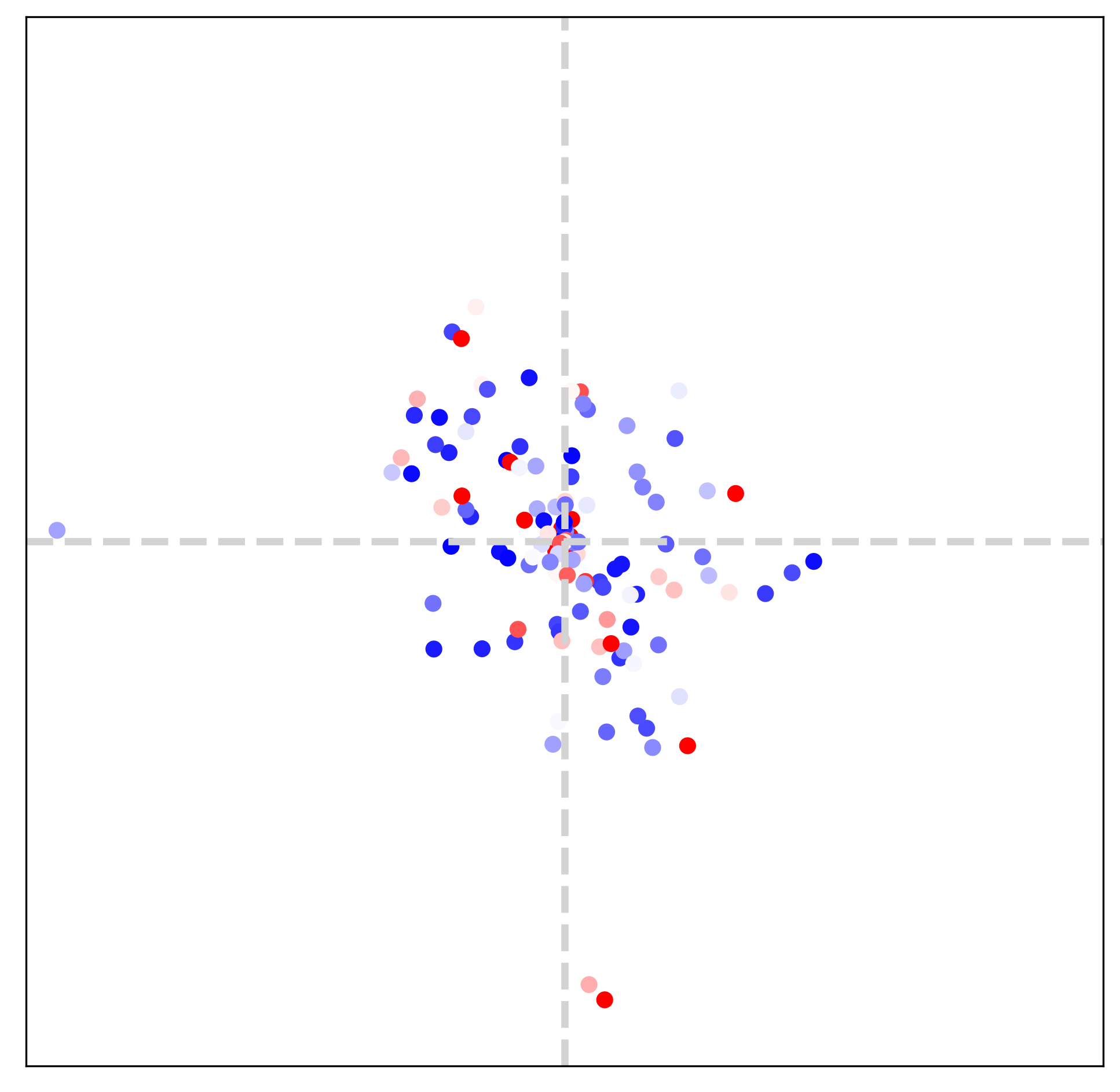}}{Diagonal}

\vspace*{-6pt}
\caption{GT image (top), GT image spectrum (middle), and corresponding estimated Fourier space from LTE (bottom) of \underline{\textbf{various textures}}. EDSR-baseline \cite{Lim_2017_CVPR_Workshops} is used as an encoder.}
\label{fig:four_vis}
\end{figure}

\subsection{Fourier Space}
\label{sec:fourier}
\begin{figure}[t]
\footnotesize
\centering
\stackunder[2pt]{\includegraphics[height=0.60in]{figure/F_map/Baseline/urban100-12-2-360-660.png}}{LTE}
\stackunder[2pt]{\includegraphics[height=0.60in]{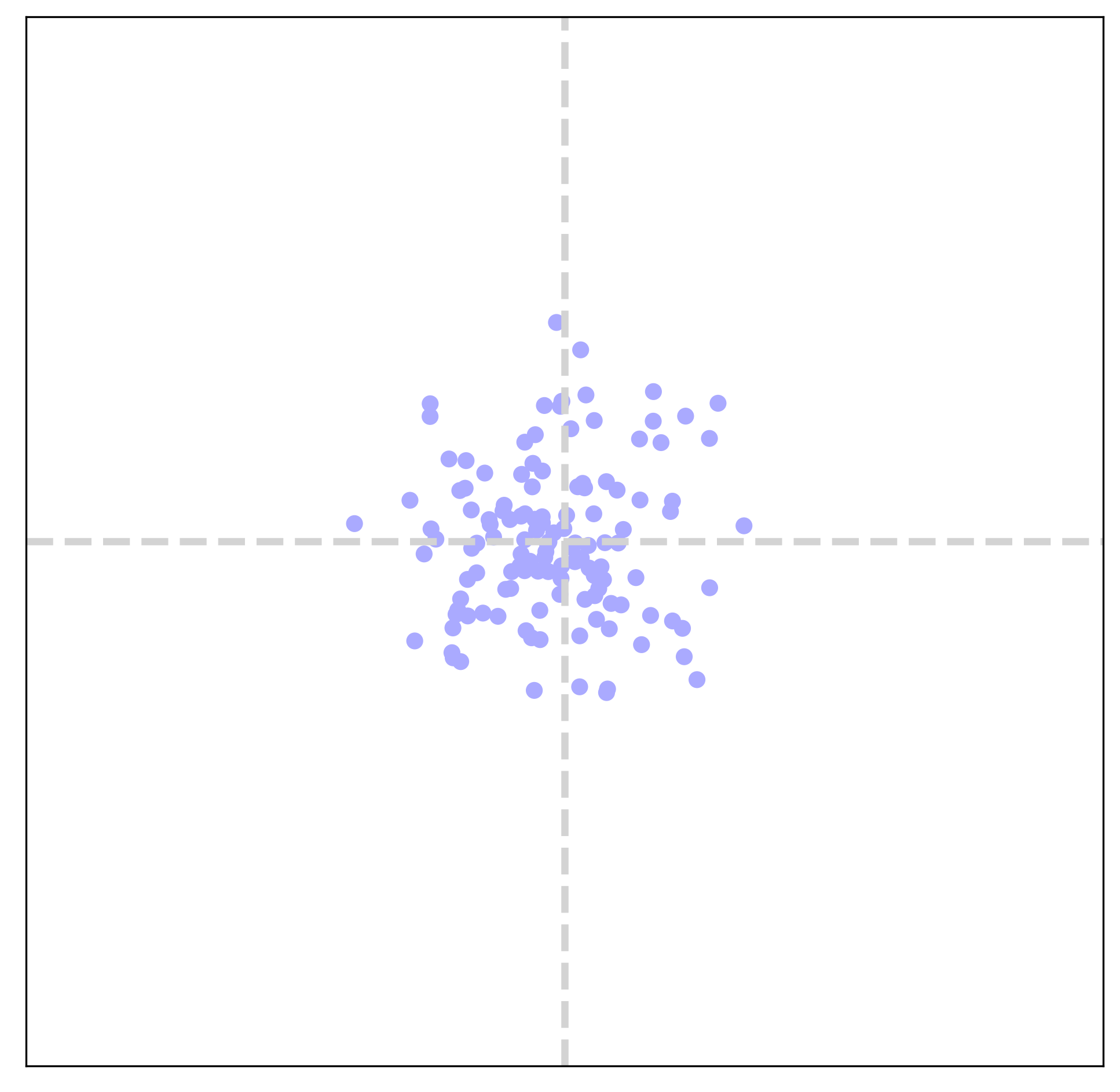}}{LTE (-A)}
\stackunder[2pt]{\includegraphics[height=0.60in]{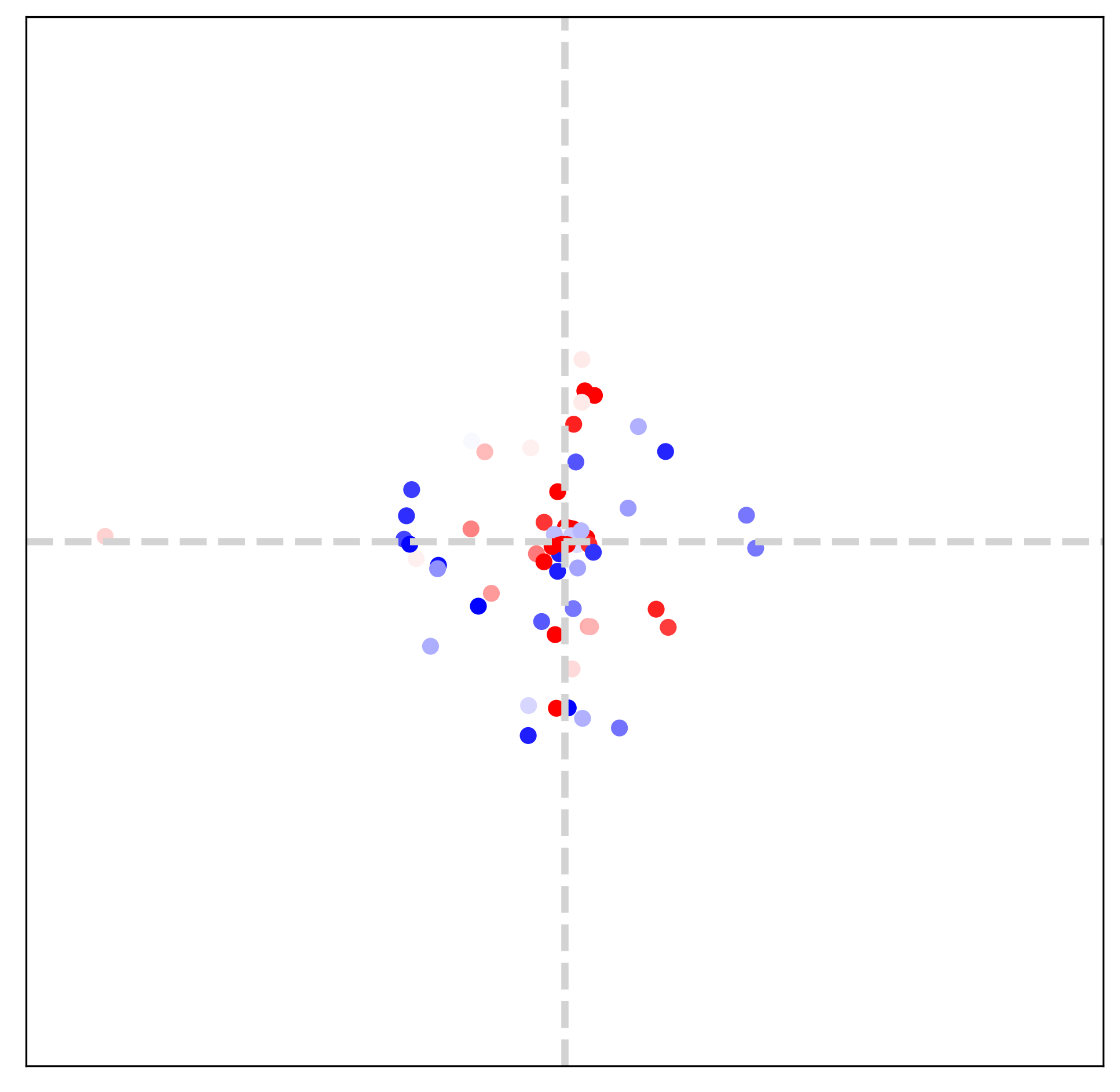}}{LTE (-F)}
\stackunder[2pt]{\includegraphics[height=0.60in]{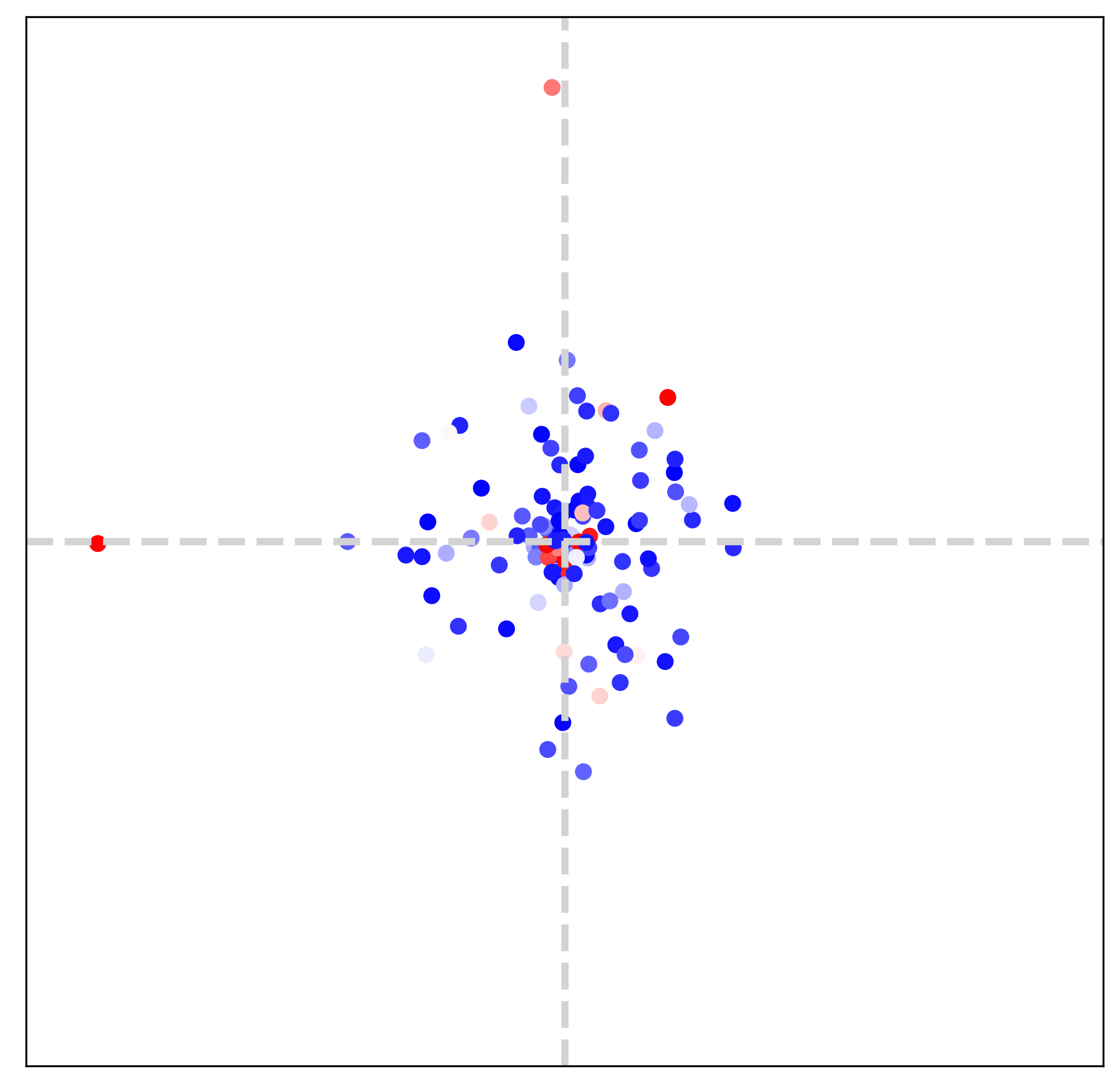}}{LTE (-P)}
\stackunder[2pt]{\includegraphics[height=0.60in]{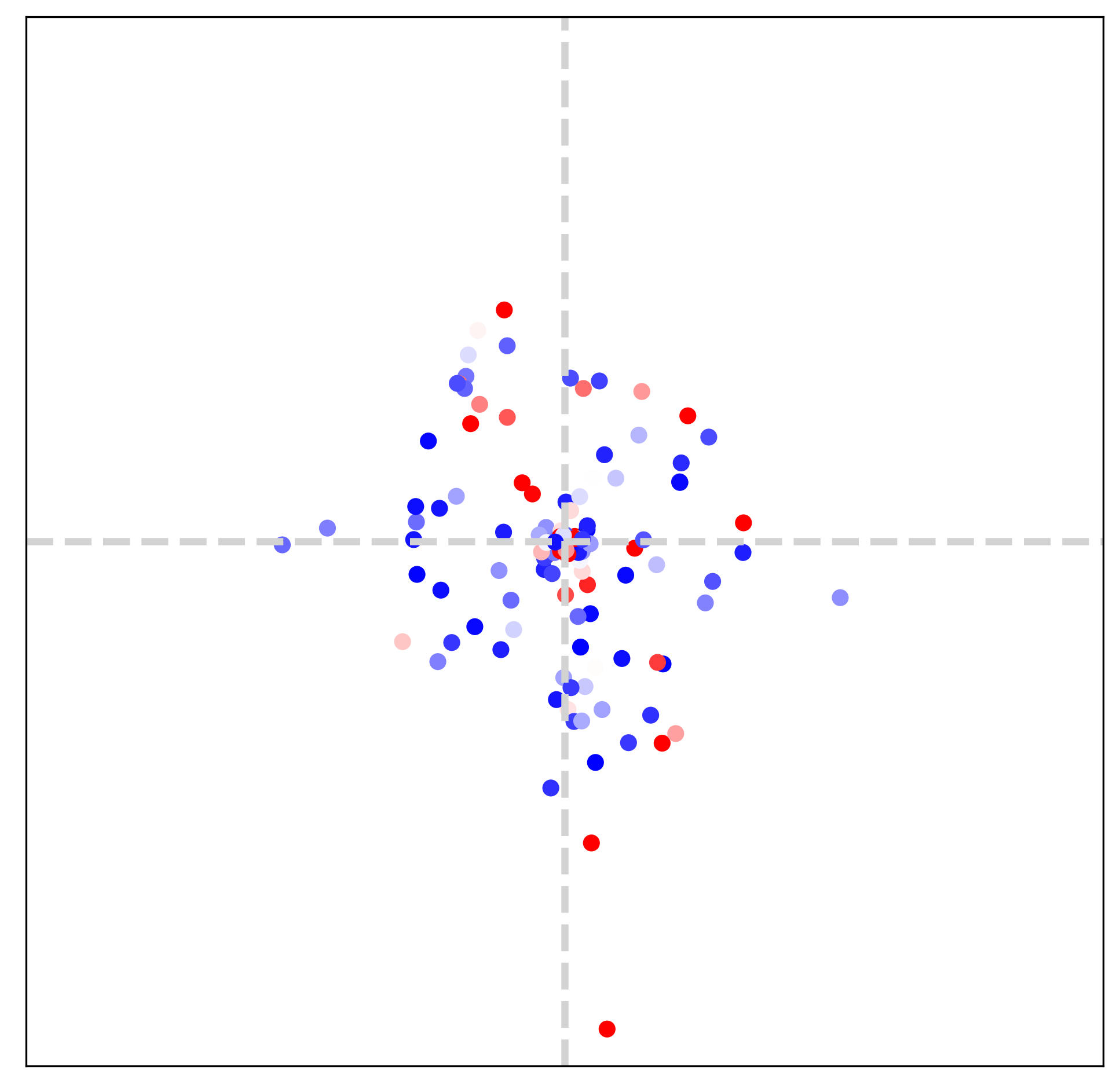}}{LTE (-L)}
\includegraphics[height=0.60in]{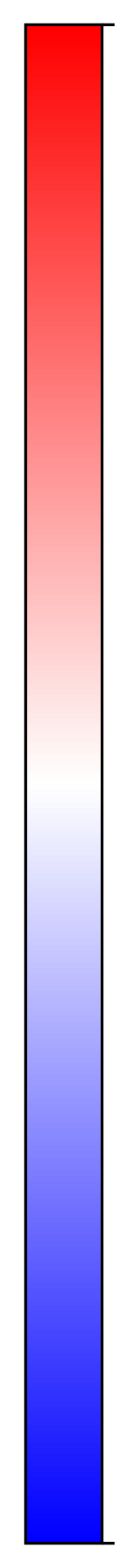}
\vspace*{-6pt}
\caption{\textbf{Ablation study and corresponding Fourier space.} -F refers to reducing the number of estimated frequencies, and -A, -P, -L refers to removing amplitude estimator, phase estimator, LR skip connection, respectively. The diagonal texture in \cref{fig:four_vis} is chosen for an ablation study, and EDSR-baseline \cite{Lim_2017_CVPR_Workshops} is used as an encoder.}
\label{fig:four_vis_abl}
\end{figure}

\begin{figure}[t]
\vspace{-6pt}
\footnotesize
\centering

\includegraphics[width=0.522in]{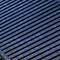}
\includegraphics[width=0.522in]{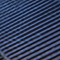}
\includegraphics[width=0.522in]{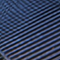}
\includegraphics[width=0.522in]{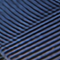}
\includegraphics[width=0.522in]{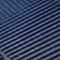}
\includegraphics[width=0.522in]{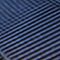}

\stackunder[2pt]{\includegraphics[width=0.522in]{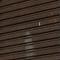}}{GT}
\stackunder[2pt]{\includegraphics[width=0.522in]{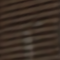}}{LTE}
\stackunder[2pt]{\includegraphics[width=0.522in]{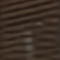}}{LTE(-A)}
\stackunder[2pt]{\includegraphics[width=0.522in]{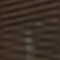}}{LTE(-F)}
\stackunder[2pt]{\includegraphics[width=0.522in]{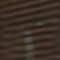}}{LTE(-P)}
\stackunder[2pt]{\includegraphics[width=0.522in]{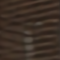}}{LTE(-L)}
\vspace*{-6pt}
\caption{\textbf{Qualitative ablation study of LTE on Urban100.} Definitions of -A, -F, -P, -L are shown in a caption of \cref{fig:four_vis_abl}. EDSR-baseline \cite{Lim_2017_CVPR_Workshops} is used as an encoder.}
\label{fig:qual_abl}
\end{figure}

In this section (\cref{fig:four_vis,fig:four:enc,fig:four_vis_abl}), we visualize extracted dominant frequencies with various textures. Furthermore, we investigate the contributions of each LTE component (specifically amplitude, frequency, phase, LR skip connection) through the lens of Fourier space.

\textbf{Setup} For visualization, we observe outputs of an amplitude estimator ($h_a$) and a frequency estimator ($h_f$). We first scatter dominant frequencies on 2D space and set color for each point with a magnitude. All scatter maps are defined on $[-1.5,1.5]^2$, and the value range of each map is different from the other. In addition, 16-tap discrete Fourier transform (DFT) of GT images are provided to compare dominant frequencies of LTE and those of GT.

\begin{table}[t]
\vspace{-6pt}
\centering
\setlength{\tabcolsep}{1.2pt}
\scriptsize
\begin{tabular}{c
|>{\centering\arraybackslash}p{0.80cm}>{\centering\arraybackslash}p{0.80cm}>{\centering\arraybackslash}p{0.80cm}
|>{\centering\arraybackslash}p{0.80cm}>{\centering\arraybackslash}p{0.80cm}
}
 & \multicolumn{3}{c|}{In-scale} & \multicolumn{2}{c}{Out-of-scale} \\
& $\times2$ & $\times3$ & $\times4$ & $\times6$ & $\times8$ \\
\hline\hline
LTE & \textbf{33.72} & \textbf{30.37} & 28.65 & \textbf{26.50} & \textbf{24.99} \\
LTE (-A) & 33.66 & 30.07 & \textbf{28.66} & 26.49  & 24.93 \\
LTE (-F) & 33.66 & 30.35 & 28.64 & \textbf{26.50}  & 24.98 \\
LTE (-P) & 33.58 & 30.26 & 28.58 & 26.40  & 24.90 \\
LTE (-L) & 33.62 & 30.36 & 28.64 & 26.48  & 24.98 \\
\end{tabular}
\vspace*{-6pt}
\caption{\textbf{Quantitative ablation study of LTE on Set14.} Definitions of -A, -F, -P, -L are shown in a caption of \cref{fig:four_vis_abl}. EDSR-baseline \cite{Lim_2017_CVPR_Workshops} is used as an encoder.}
\label{tab:Quan_abl}
\vspace{-15pt}
\end{table}

\textbf{Image texture and Fourier space} We choose three different textures: horizontal, vertical, diagonal textures, as illustrated in \cref{fig:four_vis}. Frequency maps from LTE in the bottom row are obtained from two-fold downsampled images. By comparing the middle row and the bottom row in \cref{fig:four_vis}, we observe that estimated dominant frequencies follow the dominant frequencies of GT. It indicates that LTE obtains dominant frequencies and corresponding Fourier coefficients by observing pixels inside an RF. Note that the size of RF is determined by deep SR encoders ($E_\varphi$), such as EDSR-baseline \cite{Lim_2017_CVPR_Workshops}, RDN \cite{zhang2018residual}, and SwinIR \cite{liang2021swinir}.

\textbf{Encoder and Fourier space} \cref{tab:Quan_Bench} and \cref{fig:enc} demonstrate that LTE accomplishes better performance when SwinIR \cite{liang2021swinir} is used as an encoder. \cref{fig:four:enc} supports the observation by visualizing Fourier space. We remark that SwinIR-LTE captures dominant frequencies on a diagonal axis while EDSR-baseline-LTE estimates only low-frequency components. From this observation,  LTE with a powerful encoder extracts precise dominant frequencies.

\textbf{Ablation study on Fourier space} \cref{fig:four_vis_abl} shows Fourier spaces of LTE when each component is missing. We choose a diagonal texture in \cref{fig:four_vis} for an ablation study. Please see a caption of \cref{fig:four_vis_abl} for definitions of -A, -F, -P, -L. LTE (-A) considers coefficients of all frequencies as equal since Fourier coefficients are not given from LTE. Therefore, LTE (-A) focuses on learning low-frequency content. LTE (-P) is incapable of estimating frequencies positioned on a diagonal axis. We suppose that without a scale-dependent phase estimation, LTE (-P) detects only scale-independent information: Image signal is compactly supported by low-frequency regions. We validate that the deficiency of dominant frequencies fails to learn high-frequency details by comparing LTE and LTE (-F). On comparing LTE and LTE (-L), LTE (-L) is poor in capturing dominant frequencies.

\section{Discussion}
\label{sec:dis}
\begingroup
\setlength{\columnsep}{10pt}
\setlength\intextsep{0pt}
\begin{wrapfigure}{r}{1.08in}
\centering
\includegraphics[height=0.51in]{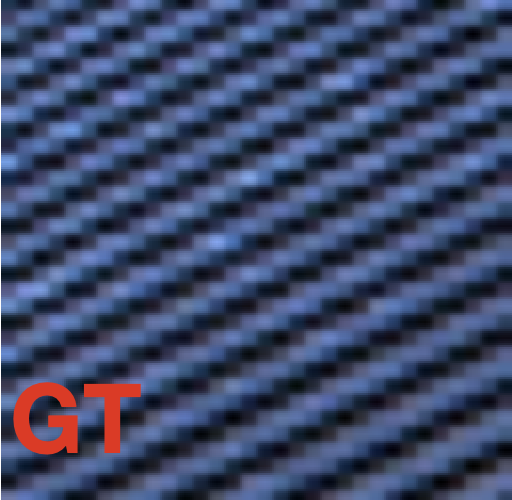}
\includegraphics[height=0.51in]{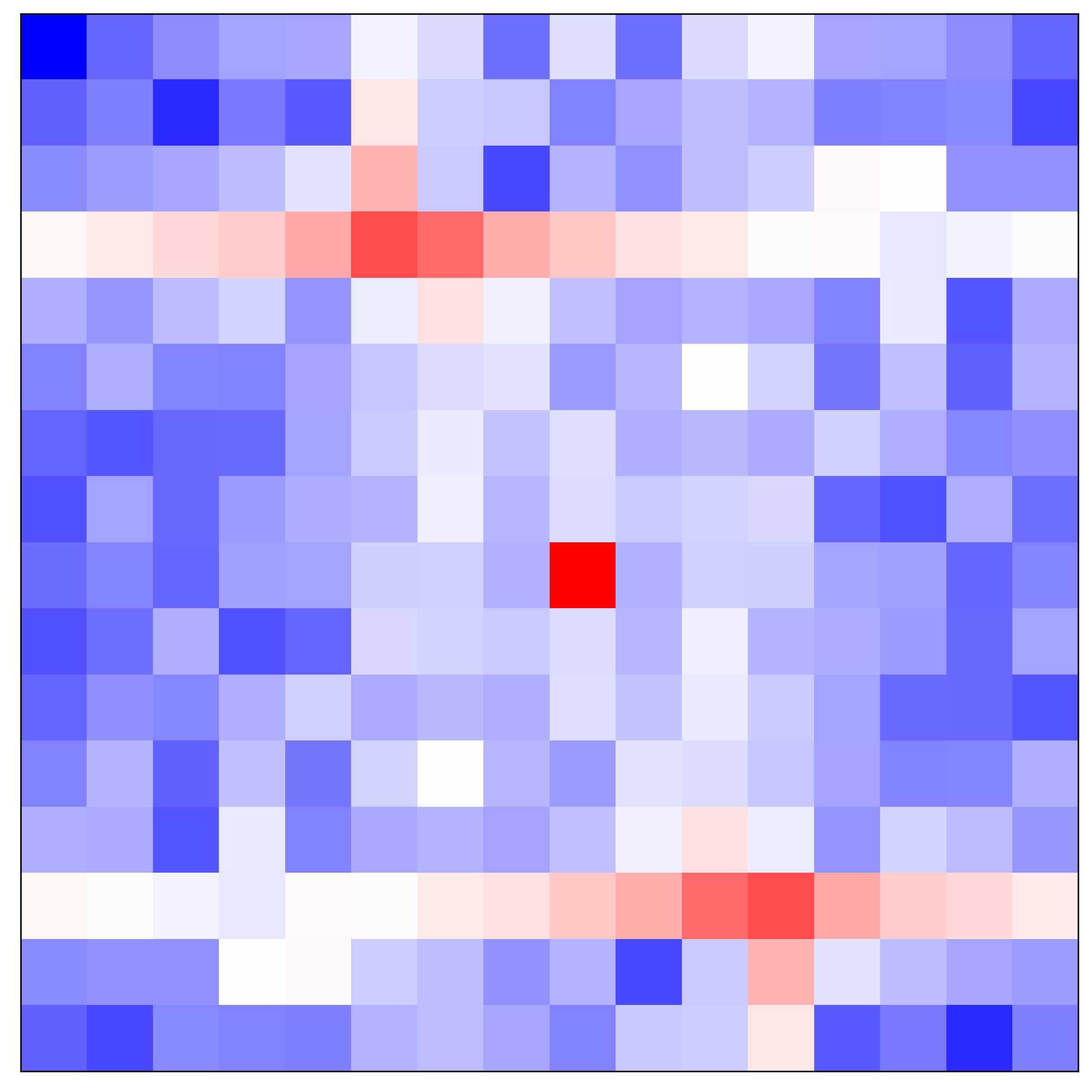}
\includegraphics[height=0.51in]{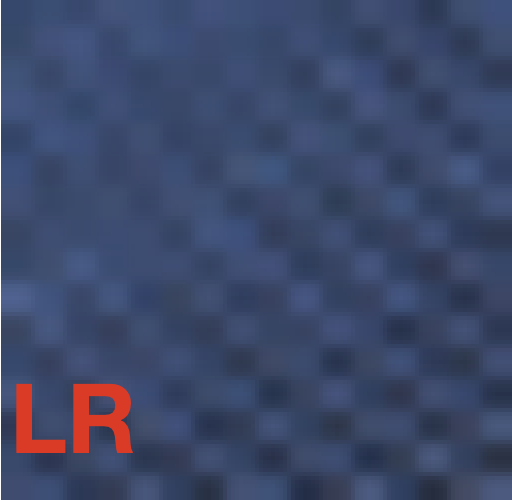}
\includegraphics[height=0.51in]{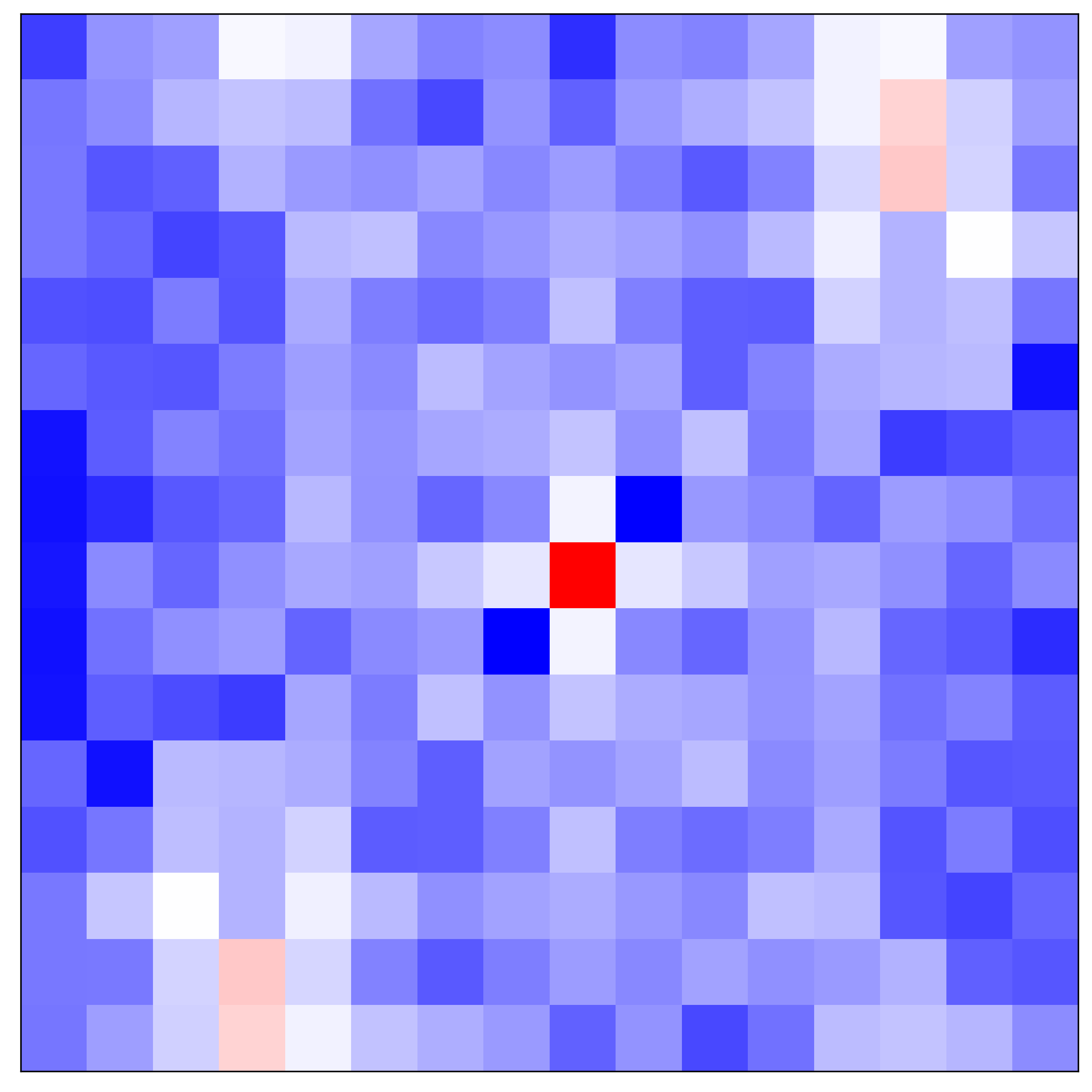}
\includegraphics[height=0.51in]{figure/LTE_design/LR_img/LR.png}
\includegraphics[height=0.51in]{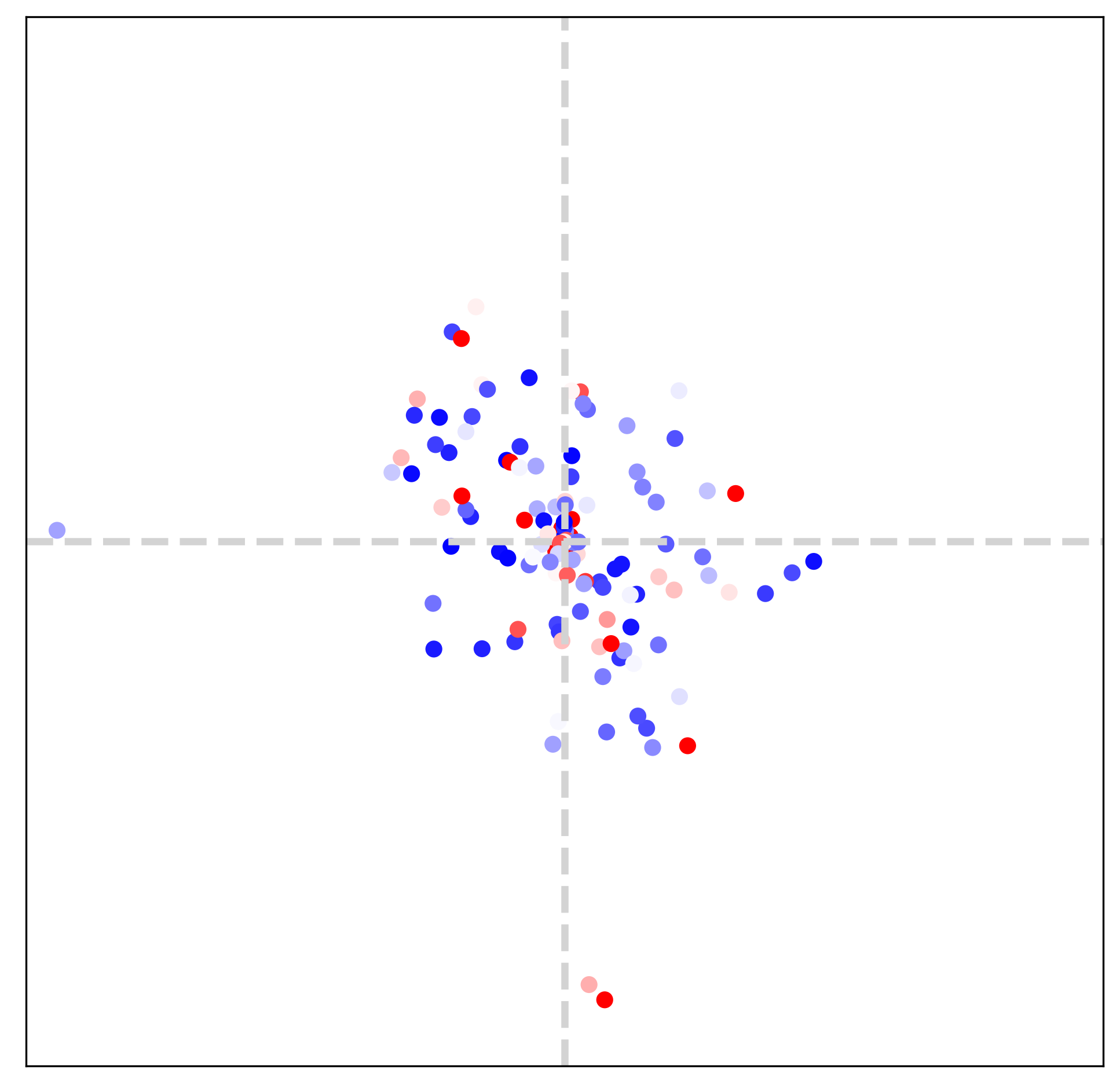}

\vspace{-6pt}
\caption{DFT and LTE}
\label{fig:LTE_design}
\end{wrapfigure}
\textbf{Advantage of LTE over DFT} In DFT, Fourier information is represented with a linear combination of image intensity. However, the DFT of an LR image with aliasing (middle) is limited in capturing dominant frequencies. In contrast, LTE with a deep neural encoder (bottom), which is a multi-chain of linear combinations and non-linear activations, is capable of estimating accurate Fourier information for an HR image (top).

\endgroup

\begingroup
\setlength{\columnsep}{10pt}
\setlength\intextsep{0pt}
\begin{wrapfigure}{r}{1.24in}
\centering
\includegraphics[height=0.51in]{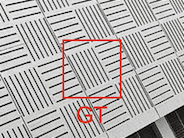}
\includegraphics[height=0.51in]{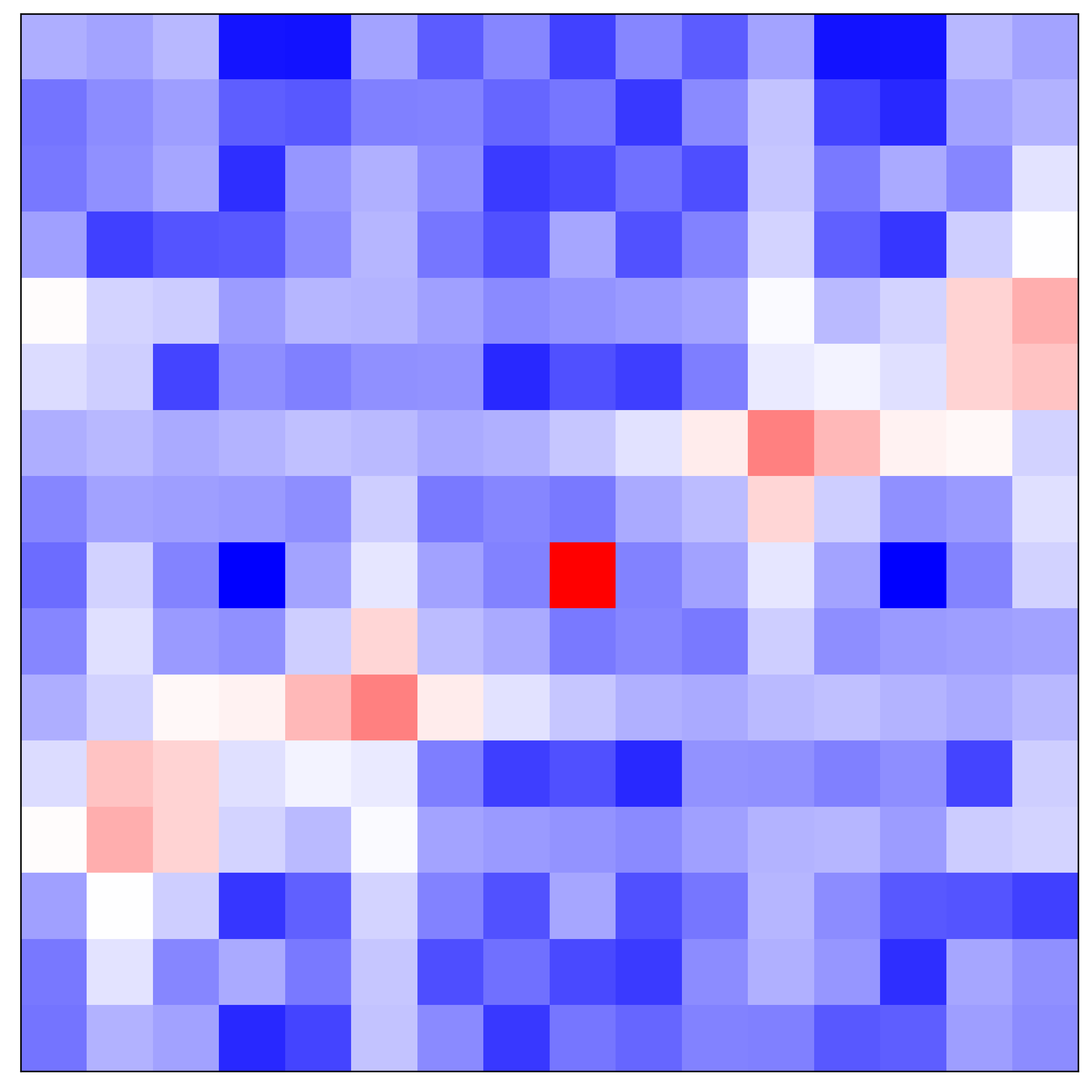}
\includegraphics[height=0.51in]{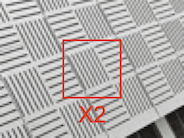}
\includegraphics[height=0.51in]{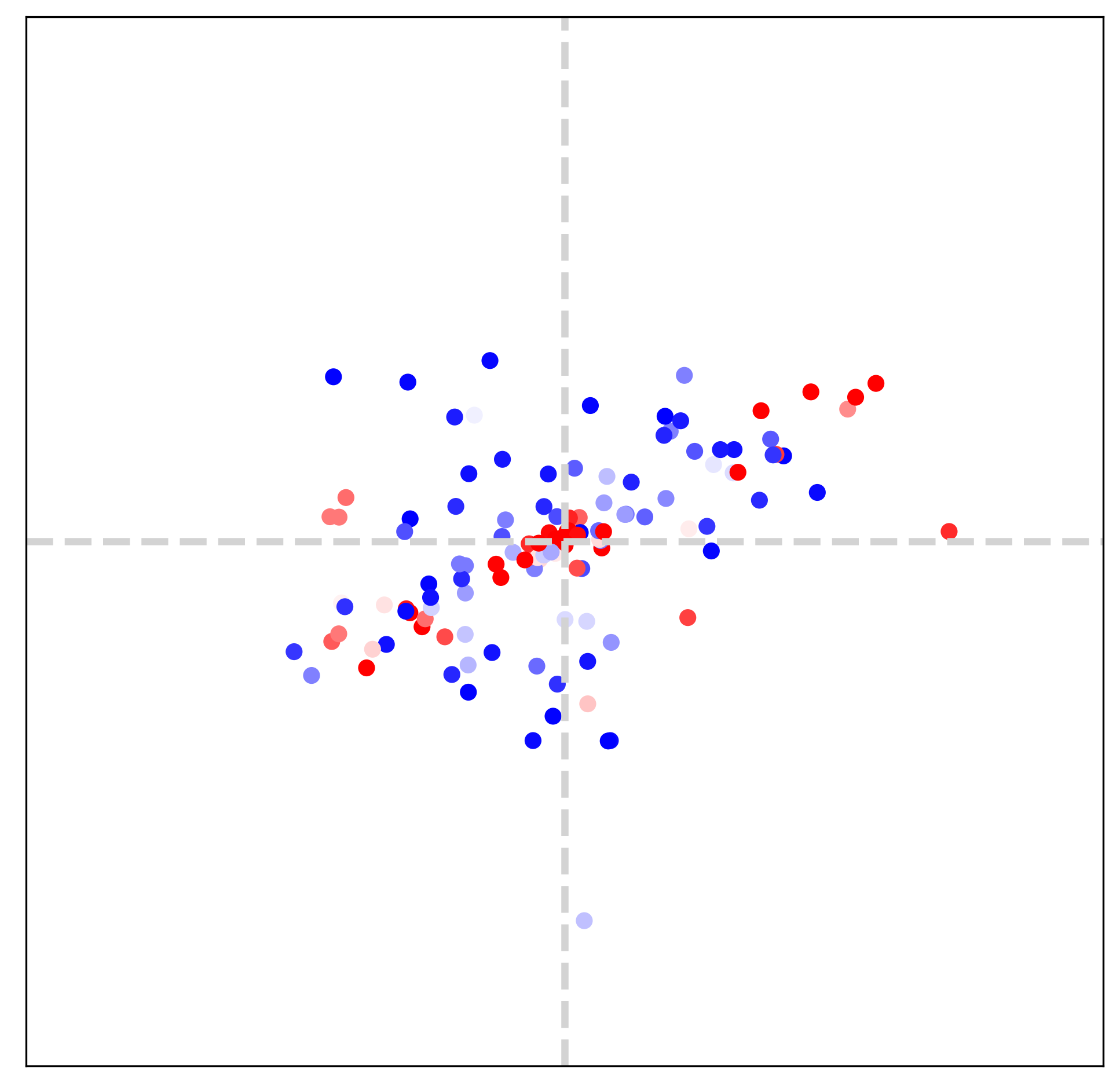}
\includegraphics[height=0.51in]{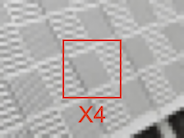}
\includegraphics[height=0.51in]{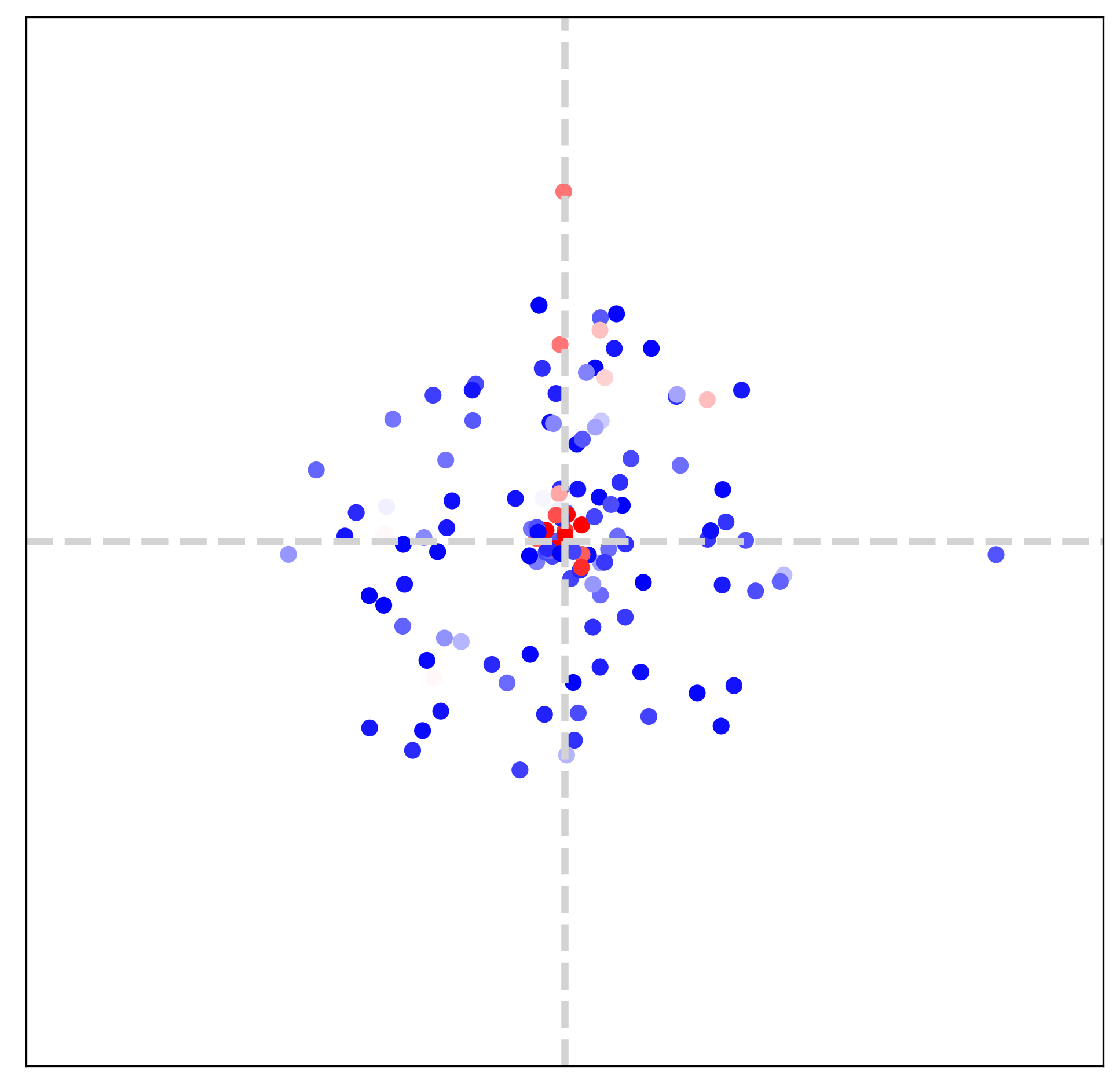}

\vspace{-6pt}
\caption{Aliasing.}
\label{fig:fail_aliasing}
\end{wrapfigure}
\textbf{Effect of aliasing} \cref{fig:four:enc} shows that SwinIR-LTE is capable of estimating dominant frequencies of natural images. In addition, the middle row in \cref{fig:fail_aliasing}  demonstrate that SwinIR-LTE extracts essential Fourier information under mild aliasing. However, such capability of SwinIR-LTE is limited when an LR image has severe aliasing. From \cref{fig:fail_aliasing}, we observe that dominant frequencies (bottom right) are inconsistent with a GT spectrum (top right) when harsh aliasing artifacts occur in an LR image (bottom left). We can resolve such limitations by extending the size of an encoder's RF, followed by increased computation and memory costs. Cost-effective architectures achieving robust performance even under severe aliasing will be investigated in future work.

\endgroup

\begingroup
\setlength{\columnsep}{10pt}
\setlength\intextsep{0pt}
\begin{wrapfigure}{r}{1.06in}
\centering
\includegraphics[width=0.51in]{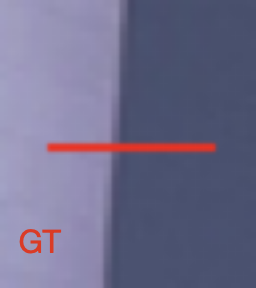}
\includegraphics[width=0.51in]{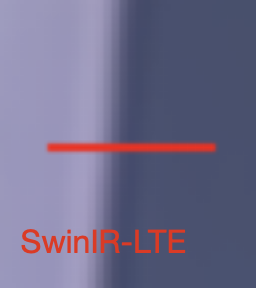}
\includegraphics[width=1.06in]{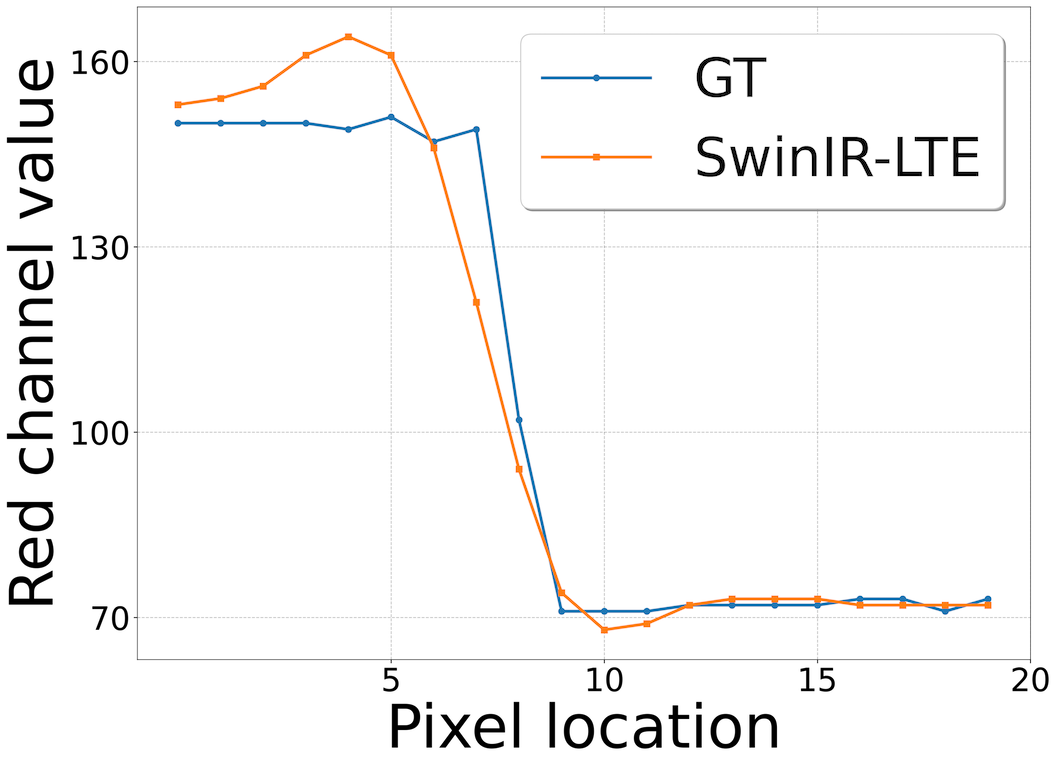}

\vspace{-6pt}
\caption{Gibbs phenomenon ($\times12$).}
\label{fig:fail}
\end{wrapfigure}
\textbf{Gibbs phenomenon} When representing continuous signals with a finite sum of Fourier basis, function overshoots at discontinuities. Such observation is referred to in the literature as the \textit{Gibbs phenomenon} or ringing artifacts in 2D images. From \cref{fig:fail}, we notice that LTE might cause overshoot at large scale factors, e.g., $\times12$. Further investigation of smoothing algorithms to alleviate such an issue is a promising direction for future work.

\endgroup

\textbf{Computation time} In practice, SR applications require short computation time. Moreover, reconstructing high-quality images, such as DIV2K, consumes extensive memory during evaluation. \cref{tab:model_complexity} compares the computation time of our LTE to other arbitrary-scale SR methods for both cases: memory-limited (top rows) and memory-consuming (bottom rows) on NVIDIA RTX 3090 24GB. To evaluate an HR image under a memory-limited condition, we compute $96\times96$ output pixels per query \cite{chen2021learning}. From the top rows of \cref{tab:model_complexity}, we observe that our LTE takes the shortest computation time while increasing memory usage. LTE has 4-layer MLP for querying, two convolution layers for estimators, and LIIF has 5-layer MLP only for querying. When querying evaluation points only once, estimators become more dominant than querying, resulting in more computation time, as in the bottom rows of \cref{tab:model_complexity}. To overcome such a limitation, we design an LTE+, which utilizes $1\times1$ convolution instead of a shared MLP for decoder implementation. Since $1\times1$ convolution has a GPU-friendly data structure, our LTE+ takes a shorter computation time and consumes less memory compared to previous works when all output pixels are queried at once.

\begin{table}[t]
\centering
\scriptsize
\begin{tabular}{c|c|c|c|c}
 \#Eval/Query & Method & \# Params. & Mem. (GB) & Time (ms) \\
\hline\hline
\multirow{3}{*}{\shortstack{9216\\($96\times 96$)}} & MetaSR \cite{hu2019meta} & 1.7M & 1.9 & 3462 \\
& LIIF \cite{chen2021learning} & 1.6M & 1.9 & 4559 \\
& LTE (ours) & 1.7M & 2.3 & 2912 \\
\hline
\multirow{4}{*}{\shortstack{1.6M\\($1248\times 1248$)}} & MetaSR \cite{hu2019meta} & 1.7M & \textit{OOM} & - \\
& LIIF \cite{chen2021learning} & 1.6M & 11.4 & 873 \\
& LTE (ours) & 1.7M & 10.2 & 925 \\
\cline{2-5}
& LTE+ (ours) & 1.7M & 7.1 & 483 \\
\end{tabular}
\vspace*{-6pt}
\caption{Memory consumption (GB) and computation time (ms) comparison to other \underline{\textbf{arbitrary-scale SR}} methods for an $\times2$ SR task. We use a $624\times624$ sized input and EDSR-baseline \cite{Lim_2017_CVPR_Workshops} as an encoder. \textit{OOM} denotes an \textit{out-of-memory}, and our GPU memory was deficient in evaluating MetaSR \cite{hu2019meta}.}
\label{tab:model_complexity}
\vspace{-15pt}
\end{table}

\section{Conclusion}
In this paper, we proposed the Local Texture Estimator (LTE) to overcome the spectral bias problem of an implicit neural function. Our LTE-based arbitrary-scale SR method consists of three components: (1) Deep SR encoder (2) LTE (3) Implicit representation function. Firstly, a deep SR encoder extracts feature maps whose height and width are the same as an LR image. Then, LTE takes feature maps from the encoder and estimates dominant frequencies with corresponding Fourier coefficients for natural images. Scale-dependent phase and LR skip connection are further provided to allow LTE to be biased in learning high-frequency textures. Finally, the implicit function reconstructs an image in arbitrary resolution using estimated Fourier information. We showed that our LTE-based neural function outperforms other arbitrary-scale SR methods in performance and visual quality with the shortest computation time.

\vspace{8pt}
\noindent
\small{\textbf{Acknowledgement} This work was partly supported by the National Research Foundation of Korea (NRF) grant funded by the Korea government (MSIT) (No. 2021R1F1A1045516, No. 2021R1A4A1028652) and Institute of Information \& communications Technology Planning \& Evaluation (IITP) grant funded by the Korea government (MSIT) (No. IITP-2021-0-02068).}



{\small
\bibliographystyle{ieee_fullname}

\begin{thebibliography}{10}\itemsep=-1pt

\bibitem{8014884}
Eirikur Agustsson and Radu Timofte.
\newblock {NTIRE 2017 Challenge on Single Image Super-Resolution: Dataset and
  Study}.
\newblock In {\em Proceedings of the IEEE Conference on Computer Vision and
  Pattern Recognition (CVPR) Workshops}, July 2017.

\bibitem{bevilacqua2012low}
Marco Bevilacqua, Aline Roumy, Christine Guillemot, and Marie line
  Alberi~Morel.
\newblock {Low-Complexity Single-Image Super-Resolution based on Nonnegative
  Neighbor Embedding}.
\newblock In {\em Proceedings of the British Machine Vision Conference}, pages
  135.1--135.10. BMVA Press, 2012.

\bibitem{DBLP:conf/cvpr/Chen000DLMX0021}
Hanting Chen, Yunhe Wang, Tianyu Guo, Chang Xu, Yiping Deng, Zhenhua Liu, Siwei
  Ma, Chunjing Xu, Chao Xu, and Wen Gao.
\newblock {Pre-Trained Image Processing Transformer}.
\newblock In {\em Proceedings of the IEEE/CVF Conference on Computer Vision and
  Pattern Recognition (CVPR)}, pages 12299--12310, June 2021.

\bibitem{chen2021learning}
Yinbo Chen, Sifei Liu, and Xiaolong Wang.
\newblock {Learning Continuous Image Representation With Local Implicit Image
  Function}.
\newblock In {\em Proceedings of the IEEE/CVF Conference on Computer Vision and
  Pattern Recognition (CVPR)}, pages 8628--8638, June 2021.

\bibitem{dai2019second}
Tao Dai, Jianrui Cai, Yongbing Zhang, Shu-Tao Xia, and Lei Zhang.
\newblock {Second-Order Attention Network for Single Image Super-Resolution}.
\newblock In {\em Proceedings of the IEEE/CVF Conference on Computer Vision and
  Pattern Recognition (CVPR)}, June 2019.

\bibitem{DBLP:conf/iclr/DosovitskiyB0WZ21}
Alexey Dosovitskiy, Lucas Beyer, Alexander Kolesnikov, Dirk Weissenborn,
  Xiaohua Zhai, Thomas Unterthiner, Mostafa Dehghani, Matthias Minderer, Georg
  Heigold, Sylvain Gelly, Jakob Uszkoreit, and Neil Houlsby.
\newblock {An Image is Worth 16x16 Words: Transformers for Image Recognition at
  Scale}.
\newblock In {\em 9th International Conference on Learning Representations,
  {ICLR} 2021, Virtual Event, Austria, May 3-7, 2021}. OpenReview.net, 2021.

\bibitem{DBLP:conf/iclr/HaDL17}
David Ha, Andrew~M. Dai, and Quoc~V. Le.
\newblock {HyperNetworks}.
\newblock In {\em 5th International Conference on Learning Representations,
  {ICLR} 2017, Toulon, France, April 24-26, 2017, Conference Track
  Proceedings}. OpenReview.net, 2017.

\bibitem{10.5555/70405.70408}
K. Hornik, M. Stinchcombe, and H. White.
\newblock Multilayer {F}eedforward {N}etworks {A}re {U}niversal
  {A}pproximators.
\newblock {\em Neural Netw.}, 2(5):359–366, July 1989.

\bibitem{hu2019meta}
Xuecai Hu, Haoyuan Mu, Xiangyu Zhang, Zilei Wang, Tieniu Tan, and Jian Sun.
\newblock {Meta-SR: A Magnification-Arbitrary Network for Super-Resolution}.
\newblock In {\em Proceedings of the IEEE/CVF Conference on Computer Vision and
  Pattern Recognition (CVPR)}, June 2019.

\bibitem{7299156}
Jia-Bin Huang, Abhishek Singh, and Narendra Ahuja.
\newblock {Single Image Super-Resolution From Transformed Self-Exemplars}.
\newblock In {\em Proceedings of the IEEE Conference on Computer Vision and
  Pattern Recognition (CVPR)}, June 2015.

\bibitem{Local_Implicit_Grid_CVPR20}
Chiyu~"Max" Jiang, Avneesh Sud, Ameesh Makadia, Jingwei Huang, Matthias
  Niessner, and Thomas Funkhouser.
\newblock {Local Implicit Grid Representations for 3D Scenes}.
\newblock In {\em IEEE/CVF Conference on Computer Vision and Pattern
  Recognition (CVPR)}, June 2020.

\bibitem{Kim_2016_CVPR}
Jiwon Kim, Jung~Kwon Lee, and Kyoung~Mu Lee.
\newblock {Accurate Image Super-Resolution Using Very Deep Convolutional
  Networks}.
\newblock In {\em Proceedings of the IEEE Conference on Computer Vision and
  Pattern Recognition (CVPR)}, June 2016.

\bibitem{DBLP:journals/corr/KingmaB14}
Diederik~P. Kingma and Jimmy Ba.
\newblock {Adam: {A} Method for Stochastic Optimization}.
\newblock In Yoshua Bengio and Yann LeCun, editors, {\em 3rd International
  Conference on Learning Representations, {ICLR} 2015, San Diego, CA, USA, May
  7-9, 2015, Conference Track Proceedings}, 2015.

\bibitem{liang2021swinir}
Jingyun Liang, Jiezhang Cao, Guolei Sun, Kai Zhang, Luc Van~Gool, and Radu
  Timofte.
\newblock {SwinIR: Image Restoration Using Swin Transformer}.
\newblock In {\em Proceedings of the IEEE/CVF International Conference on
  Computer Vision (ICCV) Workshops}, pages 1833--1844, October 2021.

\bibitem{Lim_2017_CVPR_Workshops}
Bee Lim, Sanghyun Son, Heewon Kim, Seungjun Nah, and Kyoung Mu~Lee.
\newblock {Enhanced Deep Residual Networks for Single Image Super-Resolution}.
\newblock In {\em Proceedings of the IEEE Conference on Computer Vision and
  Pattern Recognition (CVPR) Workshops}, July 2017.

\bibitem{liu2021swin}
Ze Liu, Yutong Lin, Yue Cao, Han Hu, Yixuan Wei, Zheng Zhang, Stephen Lin, and
  Baining Guo.
\newblock {Swin Transformer: Hierarchical Vision Transformer Using Shifted
  Windows}.
\newblock In {\em Proceedings of the IEEE/CVF International Conference on
  Computer Vision (ICCV)}, pages 10012--10022, October 2021.

\bibitem{martin2001database}
D. Martin, C. Fowlkes, D. Tal, and J. Malik.
\newblock {A database of human segmented natural images and its application to
  evaluating segmentation algorithms and measuring ecological statistics}.
\newblock In {\em Proceedings Eighth IEEE International Conference on Computer
  Vision. ICCV 2001}, volume~2, pages 416--423 vol.2, 2001.

\bibitem{Mei_2021_CVPR}
Yiqun Mei, Yuchen Fan, and Yuqian Zhou.
\newblock {Image Super-Resolution With Non-Local Sparse Attention}.
\newblock In {\em Proceedings of the IEEE/CVF Conference on Computer Vision and
  Pattern Recognition (CVPR)}, pages 3517--3526, June 2021.

\bibitem{Occupancy_Networks}
Lars Mescheder, Michael Oechsle, Michael Niemeyer, Sebastian Nowozin, and
  Andreas Geiger.
\newblock {Occupancy Networks: Learning 3D Reconstruction in Function Space}.
\newblock In {\em Proceedings of the IEEE/CVF Conference on Computer Vision and
  Pattern Recognition (CVPR)}, June 2019.

\bibitem{mildenhall2020nerf}
Ben Mildenhall, Pratul~P. Srinivasan, Matthew Tancik, Jonathan~T. Barron, Ravi
  Ramamoorthi, and Ren Ng.
\newblock {NeRF: Representing Scenes as Neural Radiance Fields for View
  Synthesis}.
\newblock In {\em Proceedings of the European Conference on Computer Vision
  (ECCV)}, August 2020.

\bibitem{Park_2019_CVPR}
Jeong~Joon Park, Peter Florence, Julian Straub, Richard Newcombe, and Steven
  Lovegrove.
\newblock {DeepSDF: Learning Continuous Signed Distance Functions for Shape
  Representation}.
\newblock In {\em Proceedings of the IEEE/CVF Conference on Computer Vision and
  Pattern Recognition (CVPR)}, June 2019.

\bibitem{paszke2019pytorch}
Adam Paszke, Sam Gross, Francisco Massa, Adam Lerer, James Bradbury, Gregory
  Chanan, Trevor Killeen, Zeming Lin, Natalia Gimelshein, Luca Antiga, Alban
  Desmaison, Andreas Kopf, Edward Yang, Zachary DeVito, Martin Raison, Alykhan
  Tejani, Sasank Chilamkurthy, Benoit Steiner, Lu Fang, Junjie Bai, and Soumith
  Chintala.
\newblock {PyTorch: An Imperative Style, High-Performance Deep Learning
  Library}.
\newblock In H. Wallach, H. Larochelle, A. Beygelzimer, F. d\textquotesingle
  Alch\'{e}-Buc, E. Fox, and R. Garnett, editors, {\em Advances in Neural
  Information Processing Systems}, volume~32. Curran Associates, Inc., 2019.

\bibitem{DBLP:conf/icml/RahamanBADLHBC19}
Nasim Rahaman, Aristide Baratin, Devansh Arpit, Felix Draxler, Min Lin, Fred
  Hamprecht, Yoshua Bengio, and Aaron Courville.
\newblock {On the Spectral Bias of Neural Networks}.
\newblock In Kamalika Chaudhuri and Ruslan Salakhutdinov, editors, {\em
  Proceedings of the 36th International Conference on Machine Learning},
  volume~97 of {\em Proceedings of Machine Learning Research}, pages
  5301--5310. PMLR, 09--15 Jun 2019.

\bibitem{DBLP:journals/corr/ShiCHTABRW16}
Wenzhe Shi, Jose Caballero, Ferenc Huszar, Johannes Totz, Andrew~P. Aitken, Rob
  Bishop, Daniel Rueckert, and Zehan Wang.
\newblock {Real-Time Single Image and Video Super-Resolution Using an Efficient
  Sub-Pixel Convolutional Neural Network}.
\newblock In {\em Proceedings of the IEEE Conference on Computer Vision and
  Pattern Recognition (CVPR)}, June 2016.

\bibitem{sitzmann2019siren}
Vincent Sitzmann, Julien Martel, Alexander Bergman, David Lindell, and Gordon
  Wetzstein.
\newblock {Implicit Neural Representations with Periodic Activation Functions}.
\newblock In H. Larochelle, M. Ranzato, R. Hadsell, M.~F. Balcan, and H. Lin,
  editors, {\em Advances in Neural Information Processing Systems}, volume~33,
  pages 7462--7473. Curran Associates, Inc., 2020.

\bibitem{sitzmann2019srns}
Vincent Sitzmann, Michael Zollhoefer, and Gordon Wetzstein.
\newblock {Scene Representation Networks: Continuous 3D-Structure-Aware Neural
  Scene Representations}.
\newblock In H. Wallach, H. Larochelle, A. Beygelzimer, F. d\textquotesingle
  Alch\'{e}-Buc, E. Fox, and R. Garnett, editors, {\em Advances in Neural
  Information Processing Systems}, volume~32. Curran Associates, Inc., 2019.

\bibitem{SRWarp}
Sanghyun Son and Kyoung~Mu Lee.
\newblock {SRWarp: Generalized Image Super-Resolution under Arbitrary
  Transformation}.
\newblock In {\em Proceedings of the IEEE/CVF Conference on Computer Vision and
  Pattern Recognition (CVPR)}, pages 7782--7791, June 2021.

\bibitem{tancik2020meta}
Matthew Tancik, Ben Mildenhall, Terrance Wang, Divi Schmidt, Pratul~P.
  Srinivasan, Jonathan~T. Barron, and Ren Ng.
\newblock {Learned Initializations for Optimizing Coordinate-Based Neural
  Representations}.
\newblock In {\em Proceedings of the IEEE/CVF Conference on Computer Vision and
  Pattern Recognition (CVPR)}, pages 2846--2855, June 2021.

\bibitem{tancik2020fourfeat}
Matthew Tancik, Pratul Srinivasan, Ben Mildenhall, Sara Fridovich-Keil, Nithin
  Raghavan, Utkarsh Singhal, Ravi Ramamoorthi, Jonathan Barron, and Ren Ng.
\newblock {Fourier Features Let Networks Learn High Frequency Functions in Low
  Dimensional Domains}.
\newblock In H. Larochelle, M. Ranzato, R. Hadsell, M.~F. Balcan, and H. Lin,
  editors, {\em Advances in Neural Information Processing Systems}, volume~33,
  pages 7537--7547. Curran Associates, Inc., 2020.

\bibitem{8014883}
Radu Timofte, Eirikur Agustsson, Luc Van~Gool, Ming-Hsuan Yang, and Lei Zhang.
\newblock {NTIRE 2017 Challenge on Single Image Super-Resolution: Methods and
  Results}.
\newblock In {\em Proceedings of the IEEE Conference on Computer Vision and
  Pattern Recognition (CVPR) Workshops}, July 2017.

\bibitem{Wang2020Learning}
Longguang Wang, Yingqian Wang, Zaiping Lin, Jungang Yang, Wei An, and Yulan
  Guo.
\newblock {Learning a Single Network for Scale-Arbitrary Super-Resolution}.
\newblock In {\em Proceedings of the IEEE/CVF International Conference on
  Computer Vision (ICCV)}, pages 4801--4810, October 2021.

\bibitem{DBLP:conf/iclr/XuZLDKJ21}
Keyulu Xu, Mozhi Zhang, Jingling Li, Simon~Shaolei Du, Ken{-}ichi
  Kawarabayashi, and Stefanie Jegelka.
\newblock {How Neural Networks Extrapolate: From Feedforward to Graph Neural
  Networks}.
\newblock In {\em 9th International Conference on Learning Representations,
  {ICLR} 2021, Virtual Event, Austria, May 3-7, 2021}. OpenReview.net, 2021.

\bibitem{zeyde2010single}
Roman Zeyde, Michael Elad, and Matan Protter.
\newblock {On Single Image Scale-Up Using Sparse-Representations}.
\newblock In Jean-Daniel Boissonnat, Patrick Chenin, Albert Cohen, Christian
  Gout, Tom Lyche, Marie-Laurence Mazure, and Larry Schumaker, editors, {\em
  Curves and Surfaces}, pages 711--730, Berlin, Heidelberg, 2012. Springer
  Berlin Heidelberg.

\bibitem{zhang2018rcan}
Yulun Zhang, Kunpeng Li, Kai Li, Lichen Wang, Bineng Zhong, and Yun Fu.
\newblock {Image Super-Resolution Using Very Deep Residual Channel Attention
  Networks}.
\newblock In {\em Proceedings of the European Conference on Computer Vision
  (ECCV)}, September 2018.

\bibitem{zhang2018residual}
Yulun Zhang, Yapeng Tian, Yu Kong, Bineng Zhong, and Yun Fu.
\newblock {Residual Dense Network for Image Super-Resolution}.
\newblock In {\em Proceedings of the IEEE Conference on Computer Vision and
  Pattern Recognition (CVPR)}, June 2018.

\end{thebibliography}

}

\end{document}